\pdfoutput=1

\documentclass[11pt]{article}

\usepackage[]{naacl2021}

\usepackage{times}
\usepackage{latexsym}

\usepackage[T1]{fontenc}

\usepackage[utf8]{inputenc}

\usepackage{microtype}

\usepackage{times}
\usepackage{latexsym}

\usepackage{amssymb}
\usepackage{pifont}
\usepackage{graphicx}
\usepackage{booktabs} 
\usepackage{subcaption}

\usepackage{colortbl}
\usepackage{multirow}
\usepackage{import}
\usepackage{cmll}
\usepackage{MnSymbol}
\usepackage{xspace}
\usepackage{paralist}
\newcommand{\datasetname}{\textsc{xformal}\xspace}
\newcommand{\gyafc}{\textsc{gyafc}\xspace}
\newcommand{\bert}{\textsc{bert}\xspace}
\newcommand{\bleu}{\textsc{bleu}\xspace}
\newcommand{\italian}{\textsc{it}\xspace}
\newcommand{\french}{\textsc{fr}\xspace}
\newcommand{\portuguese}{\textsc{br-pt}\xspace}
\newcommand{\english}{\textsc{en}\xspace}
\newcommand{\lsix}{L$6$ Yahoo! Corpus\xspace}
\newcommand{\fst}{\textsc{f}o\textsc{st}\xspace}
\newcommand{\st}{\textsc{st}\xspace}
\newcommand{\cmark}{\ding{51}}%
\newcommand{\xmark}{\ding{55}}%
\newcommand{\sudhajoel}{\textsc{rt}$18$\xspace}

\newcommand{\appendixmacro}{A.}

\newcommand{\SG}{^\filledstar}
\newcommand{\sg}{^\filledstar}
\newcommand*\rot{\rotatebox{90}}

\usepackage{arydshln}
\definecolor{red}{rgb}{0.5, 0.1, 0.3}
\definecolor{blue}{rgb}{0., 0.3, 0.9}
\definecolor{whitesmoke}{rgb}{0.91,0.91,0.91}
\definecolor{whitesmoke}{rgb}{0.93,0.93,0.93}

\definecolor{darkgreen}{rgb}{0.0, 0.42, 0.24}

\newcommand{\C}{\cellcolor{whitesmoke}}

\newcommand{\dg}[1]{\textcolor{darkgreen}{#1}}
\newcommand{\red}[1]{\textcolor{red}{#1}}
\newcommand{\blue}[1]{\textcolor{blue}{#1}}

\newcommand{\fil}{\cellcolor{whitesmoke}}
\newcommand{\pteight}{\cellcolor{darkgreen!3}}
\newcommand{\ptseven}{\cellcolor{darkgreen!4}}
\newcommand{\ptsix}{\cellcolor{darkgreen!5}}
\newcommand{\ptfive}{\cellcolor{darkgreen!6}}
\newcommand{\ptfour}{\cellcolor{darkgreen!10}}
\newcommand{\ptthree}{\cellcolor{darkgreen!15}}
\newcommand{\pttwo}{\cellcolor{darkgreen!23}}
\newcommand{\ptone}{\cellcolor{darkgreen!30}}
\newcommand{\ptzero}{\cellcolor{darkgreen!45}}

\newcommand{\iteight}{\cellcolor{blue!3}}
\newcommand{\itseven}{\cellcolor{blue!6}}

\newcommand{\itsix}{\cellcolor{blue!9}}
\newcommand{\itfive}{\cellcolor{blue!12}}
\newcommand{\itfour}{\cellcolor{blue!15}}
\newcommand{\itthree}{\cellcolor{blue!20}}
\newcommand{\ittwo}{\cellcolor{blue!23}}
\newcommand{\itone}{\cellcolor{blue!30}}
\newcommand{\itzero}{\cellcolor{blue!45}}

\newcommand{\freight}{\cellcolor{red!2}}
\newcommand{\frseven}{\cellcolor{red!5}}
\newcommand{\frsix}{\cellcolor{red!8}}
\newcommand{\frfive}{\cellcolor{red!10}}
\newcommand{\frfour}{\cellcolor{red!12}}
\newcommand{\frthree}{\cellcolor{red!15}}
\newcommand{\frtwo}{\cellcolor{red!23}}
\newcommand{\frone}{\cellcolor{red!30}}
\newcommand{\frzero}{\cellcolor{red!45}}

\usepackage{xcolor}

\newcommand{\mycomment}[3]{}

\newcommand{\ignore}[1]{}
\usepackage{microtype}

\usepackage[colorinlistoftodos]{todonotes}

\title{Olá, Bonjour, Salve! \\ \textsc{xformal}: A Benchmark for Multilingual Formality Style Transfer }

\author{
  Eleftheria Briakou\thanks{~\ Work done as a Research Intern at Dataminr Inc.} \\
  University of Maryland\\
  College Park, MD, USA\\
  \small \texttt{{ebriakou@cs.umd.edu}}\\ \And

  Di Lu, Ke Zhang, Joel Tetreault  \\
  Dataminr, Inc. \\
  New York, ND, USA\\
  \small \texttt{\{dlu,kzhang,jtetreault\}@dataminr.com}\\
 }
 
 \author{
  Eleftheria Briakou\thanks{~\ Work done as a Research Intern at Dataminr, Inc.} \\
  University of Maryland\\
  \small \texttt{\href{mailto:ebriakou@cs.umd.edu}{ebriakou@cs.umd.edu}}\\\And

  Di Lu  \\
  Dataminr, Inc. \\
  \small \texttt{\href{dlu@dataminr.com}{dlu@dataminr.com}}\\\And
  
  \textbf{Ke Zhang} \\
  Dataminr, Inc. \\
  \small \texttt{\href{mailto:kzhang@dataminr.com}{kzhang@dataminr.com}}\\\And
  
  \textbf{Joel Tetreault} \\
  Dataminr, Inc. \\
  \small \texttt{\href{mailto:jtetreault@dataminr.com}{jtetreault@dataminr.com}}}

\begin{document}
\maketitle

\begin{abstract}
   
    We take the first step towards multilingual style transfer by creating and releasing \datasetname, a benchmark of multiple formal reformulations of informal text in  Brazilian Portuguese, French, and Italian. Results on \datasetname suggest that state-of-the-art style transfer approaches perform close to simple baselines, indicating that style transfer is even more challenging when moving multilingual.\footnote{Code and data: \href{https://github.com/Elbria/xformal-FoST}{https://github.com/Elbria/xformal-FoST}} 
    
\end{abstract}

\section{Introduction}
Style Transfer (\textsc{st}) is the task of automatically transforming text in one style into another (for example, making an impolite request more polite).  Most work in this growing field has focused primarily on style transfer within English, while \textit{covering different languages} has received disproportional interest. Concretely, out of $35$ \st papers we reviewed, all of them report results for \st within English text, while there is just a single work covering each of the following languages: Chinese, Russian, Latvian, Estonian, and French \cite{shang-etal-2019-semi,tikhonov-etal-2019-style,korotkova-2019,niu-etal-2018-multi}. Notably, even though some efforts have been made towards multilingual \st, researchers are limited to providing system outputs as a means of evaluation, and progress is hampered by the scarcity of resources for most languages. 
\begin{table}[tbp]
    \centering
     \scalebox{0.85}{
    \begin{tabular}{l}
    
     \rowcolor{gray!10}
    \textsc{brazilian-portuguese} \\\addlinespace[0.3em]
    \textcolor{darkgreen}{saiam disso, força} de vontade!! \\
    \it \textcolor{gray}{get out of it, willpower!!}\\\addlinespace[0.15cm]
    \textcolor{darkgreen}{Abandonem essa situação, tenham força} \it de vontade.\\
    \it   \textcolor{gray}{Abandon this situation, have willpower!}\\
    
        \addlinespace[0.6em]

    \rowcolor{gray!10}
    \textsc{french} \\ \addlinespace[0.3em]
     Il avait \textcolor{red}{les yeux braqués ailleurs.} \\ 
      \it \textcolor{gray}{He had his eyes fixed elsewhere.}\\ \addlinespace[0.15cm]
      
      Il ne \textcolor{red}{prêtait pas attention à la situation.}\\ 
      \it \textcolor{gray}{He was not paying attention to the situation.}\\
      
    \addlinespace[0.6em]

    \rowcolor{gray!10}
    \textsc{italian} \\ \addlinespace[0.3em]   
    \textcolor{blue}{in bocca al lupo!} \\
    \it \textcolor{gray}{good luck!}\\ \addlinespace[0.15cm]
    \textcolor{blue}{Ti rivolgo un sincero augurio!}\\ 
    \it \textcolor{gray}{I send you a sincere wish!}\\

    \end{tabular}}
    \caption{Informal-Formal pairs in \datasetname.}
    \vspace{-1em}
    \label{tab:motivated_examples}
\end{table}
At the same time, \st lies at the core of human communication: when humans produce language, they condition their choice of grammatical and lexical transformations to a target audience and a specific situation. Among the many possible stylistic variations, \citet{Heylighen99formalityof} argue that \textit{``a dimension similar to \textbf{formality} appears as the most \textbf{important} and \textbf{universal} feature distinguishing styles, registers or genres in \textbf{different languages}''}. 
Consider the informal excerpts and their formal reformulations in French (\french) and Brazilian Portuguese (\portuguese) in Table~\ref{tab:motivated_examples}. Both informal-formal pairs share the same content. However, the informal language conveys more information than is contained in the literal meaning of the words~\cite{hovy87}. 
These examples relate to the notion of \textit{deep formality} \cite{Heylighen99formalityof}, where the ultimate goal is that of adding the context needed to disambiguate an expression. On the other hand, variations in formality might just reflect different situational and personal factors, as shown in the Italian (\italian) example.

This work takes the first step towards a more language-inclusive direction for the field of \st by building the first corpus of style transfer for non-English languages. In particular, we make the following contributions:
\begin{inparaenum} 
    \textbf{ \item } Building upon prior work on Formality Style Transfer (\fst) \cite{rao-tetreault-2018-dear}, we contribute an \textbf{evaluation dataset}, \datasetname that consists of multiple formal rewrites of informal sentences in three Romance languages:  Brazilian Portuguese (\portuguese), French (\french), and Italian (\italian);
    \textbf{ \item } Without assuming access to any gold-standard training data for the languages at hand, we \textbf{benchmark a myriad of leading \st baselines} through automatic and human evaluation methods.   Our results show that \fst in non-English languages is particularly challenging as complex neural models perform on par with a simple rule-based system consisting of hand-crafted transformations.
\end{inparaenum}
 We make \datasetname, our annotations protocols, and analysis code publicly available and hope that this study facilitates and encourages more research towards Multilingual \st.

\section{Related Work}
Controlling style aspects in generation tasks is studied in monolingual settings with an English-centric focus (intra-language) and cross-lingual 
settings together with Machine Translation~(\textsc{mt}) (inter-language). Our work rests in intra-language \st with a multilingual focus, in contrast to prior work. 
\paragraph{\st datasets} that consist of parallel pairs in different styles include: \gyafc for formality \cite{rao-tetreault-2018-dear}, Yelp \cite{shen-2017-nips} and Amazon Product Reviews for sentiment \cite{amazon}, political slant and gender controlled datasets \cite{prabhumoye-etal-2018-style}, Expert Style Transfer \cite{cao-etal-2020-expertise}, \textsc{pastel} for imitating personal \cite{kang19bemnlp}, \textsc{simile} for simile generation~\cite{chakrabarty-etal-2020-generating}, and others.

\paragraph{Intra-language \st} was first cast as generation task by \citet{xu-etal-2012-paraphrasing} and 
is addressed through methods that use either parallel data or unpaired corpora of different styles. \textbf{Parallel corpora designed for the task at hand} are used to train traditional encoder-decoder architectures \cite{rao-tetreault-2018-dear}, learn mappings between latent representation of different styles~\cite{shang-etal-2019-semi}, or fine-tune pre-trained models~\cite{wang-etal-2019-harnessing}. Other approaches use \textbf{parallel data from similar tasks} to facilitate transfer in the target style via domain adaptation~\cite{li-etal-2019-domain}, multi-task learning~\cite{niu-etal-2018-multi,niu2019controlling}, and zero-shot transfer~\cite{korotkova-2019} or create \textbf{pseudo-parallel data} via
 data augmentation techniques~\cite{zhang-etal-2020-parallel,krishna-etal-2020-reformulating}.
Approaches that rely on \textbf{non-parallel data} include \textit{disentanglement methods} based on the idea of learning style-agnostic latent representations (e.g., \citet{shen-2017-nips, pmlr-v70-hu17e}). However, they are recently criticized for resulting in poor content preservation~\cite{unpaired-sentiment-translation, IMat, Luo19DualRL,lample-2019-multi} and alternatively, \textit{translation-based models} are proposed that use reconstruction and back-translation losses (e.g., \citet{NIPS2018_7757, prabhumoye-etal-2018-style}). Another line of work, focuses on \textit{manipulation methods} that remove the style-specific attribute of text
~(e.g., \citet{li-etal-2018-delete, unpaired-sentiment-translation}), while recent approaches use \textit{reinforcement learning} (e.g.,~\citet{WuRLS19, gong-etal-2019-reinforcement}, probabilistic formulations~\cite{he2020a}, and masked language models~\cite{malmi-etal-2020-unsupervised}.
\paragraph{Inter-language \st} is introduced by \citet{mirkin-2015} who proposed personalized \textsc{mt} for \english-French and \english-German. Subsequent \textsc{mt} works control for politeness~\cite{sennrich-etal-2016-controlling}, voice~ \cite{yamagishi-etal-2016-controlling}, personality traits~\cite{rabinovich-etal-2017-personalized}, user-provided terminology~\cite{hasler-etal-2018-neural}, gender~\cite{vanmassenhove-etal-2018-getting}, formality~\cite{niu-etal-2017-study,feely-etal-2019-controlling}, morphological variations~\cite{moryossef-etal-2019-filling}, complexity \cite{agrawal-carpuat-2019-controlling} and reading level~\cite{marchisio-etal-2019-controlling}.
\section{\textnormal{\textsc{xformal}} Collection}\label{sec:data_collection}
We describe the process of collecting formal rewrites using data statements protocols~\cite{bender-friedman-2018-data, datasheets}.
\paragraph{Curation rational} To collect \datasetname, we firstly curate informal excerpts in multiple languages. To this end, we follow
the procedures described in~\citet{rao-tetreault-2018-dear} (henceforth \sudhajoel)  who create a corpus of informal-formal sentence-pairs in English (\english) entitled Grammarly's Yahoo Answers Formality Corpus (\gyafc).

Concretely, we use the L$6$ Yahoo! Answers corpus 
that consists of questions and answers posted to the Yahoo! Answers platform.\footnote{\url{https://webscope.sandbox.yahoo.com/catalog.php?datatype=l&did=11}}
The corpus contains a large number of informal text and allows control for different languages and different domains.\footnote{ More details are included under \appendixmacro\ref{sec:multi_stats_l6}.} Similar to the collection of \gyafc, we extract all answers from the Family \& Relationships (F\&R) topic  
that correspond to the three languages of interest: \textit{Família e Relacionamentos}~(\textsc{br-pt}), \textit{Relazioni e famiglia}~(\textsc{it }), and \textit{Amour et relations}~(\textsc{fr}) (\textbf{Step 1}).
We follow the same pre-processing steps as described in~\sudhajoel for consistency (\textbf{Step 2}). We filter out answers that: a) consist of questions; b) include \textsc{url}s; c) have fewer than five or more than $25$ tokens; or d) constitute duplicates.\footnote{We tokenize with nltk: \url{https://www.nltk.org/api/nltk.tokenize.html}} We automatically extract informal candidate sentences, as described in \S\ref{sec:automatic_eval} (\textbf{Step 3}). Finally, we randomly sample $1{,}000$ sentences from the pool of informal candidates for each language. Table~\ref{tab:l6_statistics} presents statistics of the curation steps.
\begin{table}[tpb]
    \centering
    \scalebox{0.95}{
    \begin{tabular}{lrrr}
    \rowcolor{gray!10}
     Corpus         & \portuguese   & \french  & \italian    \\

    \toprule[1.2pt]
    \addlinespace[0.5em]
    
    L$6$ Yahoo!             & $230{,}302$  & $225{,}476$ & $101{,}894$  \\
    \addlinespace[0.2em]

    \hspace{0.2em} $+$ \textbf{Step 1}         & $37{,}356$   & $34{,}849$ & $13{,}443$    \\
    \hspace{0.2em} $+$ \textbf{Step 2}         & $14{,}448$   & $14{,}856$  & $4{,}095$     \\
    \hspace{0.2em} $+$ \textbf{Step 3}         & $8{,}617$    & $11{,}118$  & $2{,}864$    
    \end{tabular}}
    \caption{Number of sentences in filtered versions of the \lsix across curation steps and languages.}
    \label{tab:l6_statistics}\vspace{-1.4em}
\end{table}
\paragraph{Procedures} We use the Amazon Mechanical Turk (\textsc{mt}urk) platform to collect formal rewrites for our informal sentences. For each language, we split the annotation into $20$ batches of $50$ Human Intelligence Tasks (\textsc{hit}s). In each \textsc{hit}, Turkers are given an informal excerpt and asked to \textit{generate its \textbf{formal rewrite} in the same language \textbf{without changing its meaning}}. We collect $4$ rewrites per excerpt and release detailed instructions under \appendixmacro\ref{sec:intructions_for_rewrites}.
\paragraph{Annotation Workflow \& Quality Control} 
Our annotation protocol consists of multiple Quality Control~(\textsc{qc}) steps
to ensure the recruitment of high-quality annotators. 
As a first step, we use \textbf{location restrictions~(\textsc{qc1})} to limit the pool of workers to countries where native speakers are most likely to be found. Next, we run several small \textit{pilot studies} (of $10$ \textsc{hit}s) to recruit potential workers. To participate in the pilot study, Turkers have to pass a \textbf{qualification test~(\textsc{qc2})} consisting of multiple-choice questions that test workers' understanding of formality~(see \appendixmacro\ref{sec:qualification_tests}).  The pilot study results are \textbf{reviewed by a native speaker~(\textsc{qc3})} of each language to exclude workers who performed consistently poorly. We find that the two main reasons for poor quality are: a) rewrites of minimum-level edits, or b) rewrites that change the input's meaning.  Table~\ref{tab:worker_statistics} presents the number of workers at each \textsc{qc} step. Only workers passing all quality control steps (last row of Table~\ref{tab:worker_statistics}) contribute to the final task. Finally, we \textbf{post-process} the collected rewrites by a) removing instances consisting of normalization-based edits only and b) correcting minor spelling errors using an off-the-shelf tool.\footnote{\url{https://languagetool.org/}} 
\begin{table}[tpb]
    \centering
    \scalebox{0.92}{
    \begin{tabular}{llrrr}

    \rowcolor{gray!10}
     Step & Description & \portuguese   & \french   & \italian    \\
     \toprule[1.2pt]
    \addlinespace[0.5em] 

    \textbf{(\textsc{qc1})} & Location restriction        &   $151$  & $78$ & $59$ \\
    \textbf{(\textsc{qc2})} & Qualification test          &   $54$   & $40$ & $33$ \\
    \textbf{(\textsc{qc3})} & Rewrites review             &   $\mathbf{9}$ & $\mathbf{16}$ & $\mathbf{11}$ \\
    \end{tabular}}
    \caption{Number of Turkers \textbf{after} each \textsc{qc} step.}
    \label{tab:worker_statistics}\vspace{-4mm}
\end{table}
\paragraph{Turkers' demographics} 
We recruit Turkers from Brazil, France/Canada, and Italy for \portuguese, \french, and \italian, respectively. Beyond their country of residence, no further information is available. 
\paragraph{Compensation} We compensate at a rate of \$$0.10$ per \textsc{hit} with
additional one-time bonuses that bumps them up to a target rate of over \$$10$/hour.
\\

\noindent
After this entire process, we have constructed a high-quality corpus of formality rewrites of $1{,}000$ sentences for three languages.  In the next section, we provide statistics and an analysis of \datasetname. 

\section{\textnormal{\textsc{xformal}} Statistics \& Analysis}\label{sec:data_analysis}
\begin{figure*}[tbp]
  \centering
    \begin{subfigure}[b]{0.22\linewidth}
        \includegraphics[width=\linewidth]{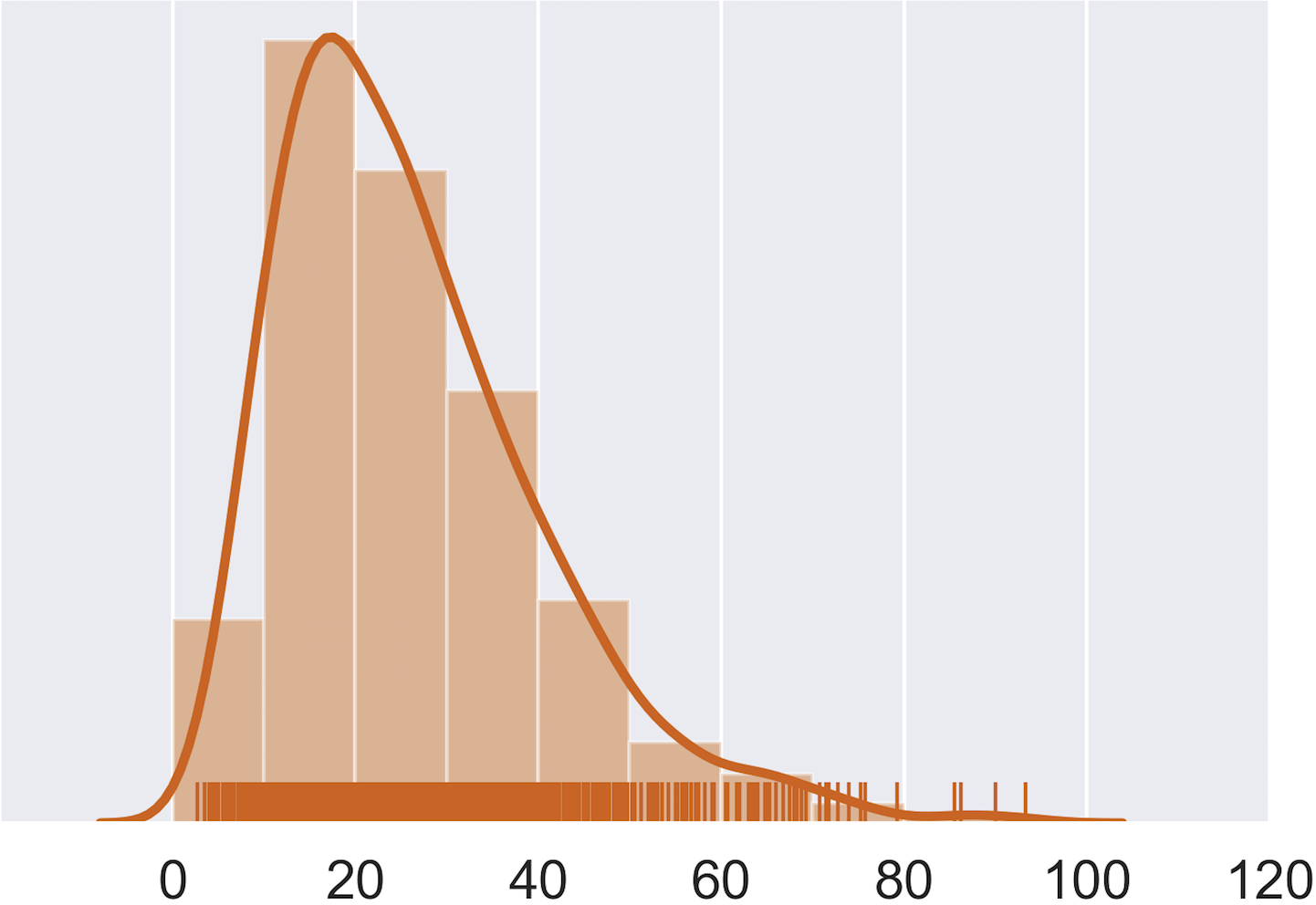}
        \caption{\portuguese}
    \end{subfigure}
    \begin{subfigure}[b]{0.22\linewidth}
        \includegraphics[width=\linewidth]{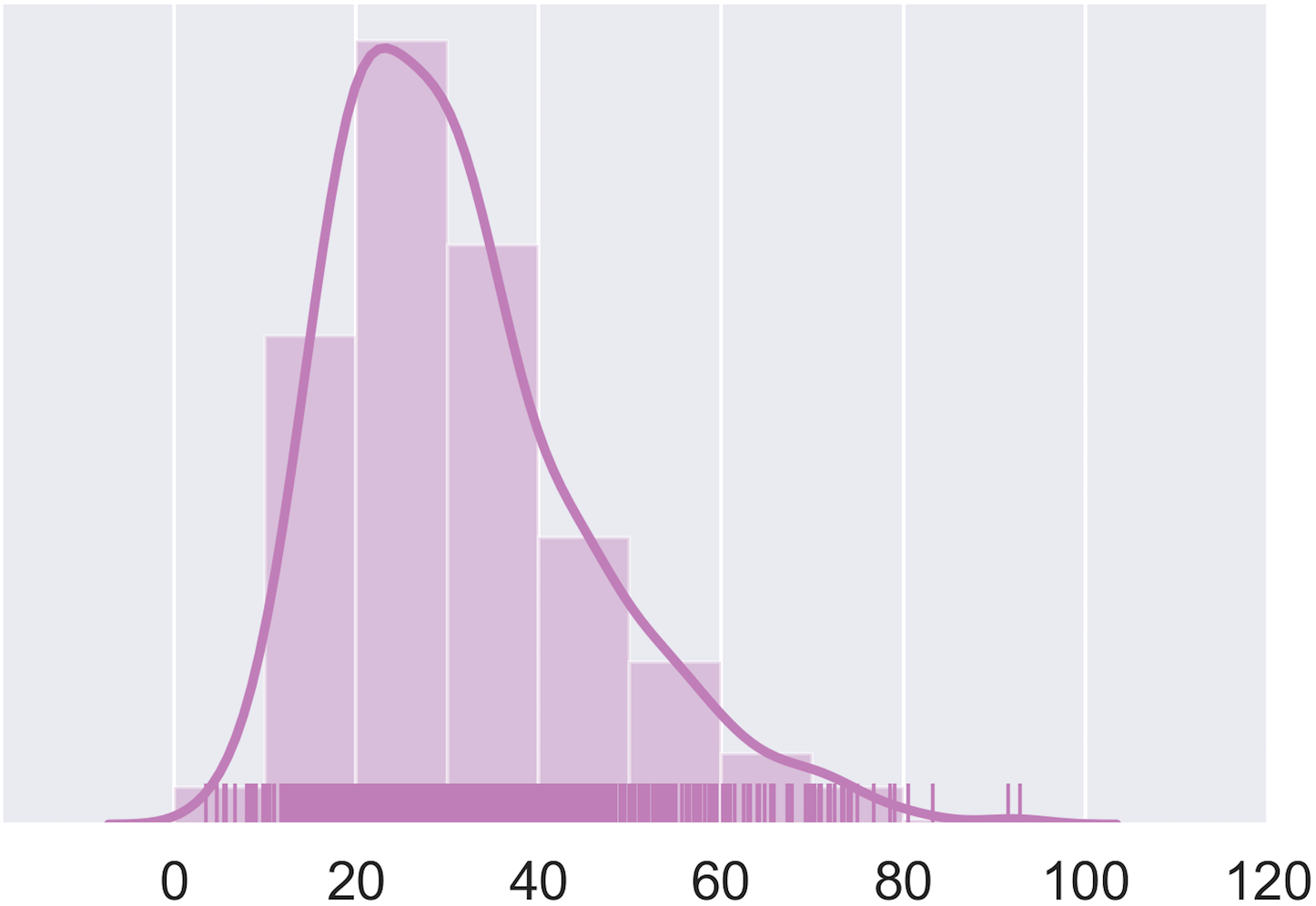}
        \caption{\french}
    \end{subfigure}
    \begin{subfigure}[b]{0.22\linewidth}
        \includegraphics[width=\linewidth]{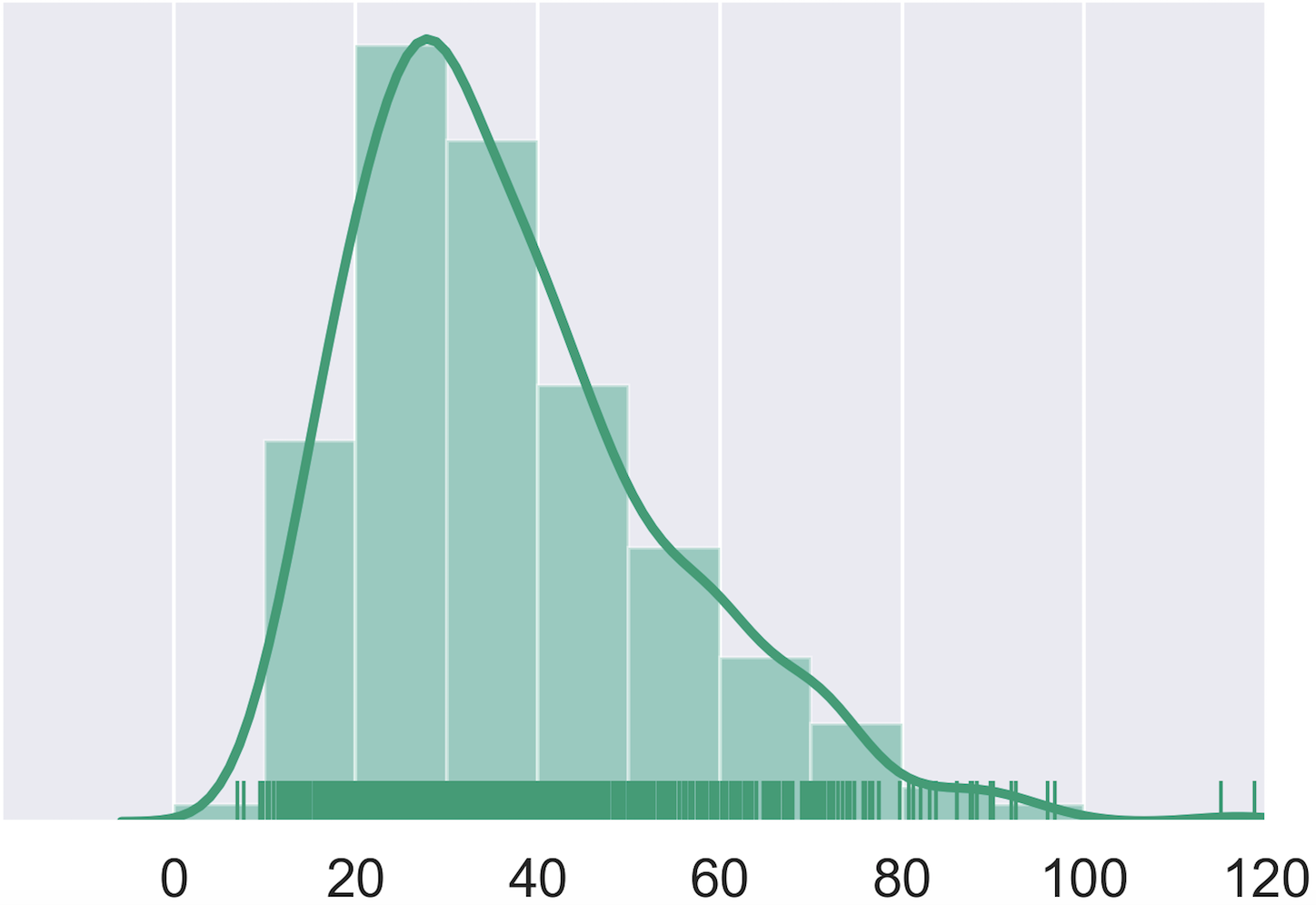}
        \caption{\italian}
    \end{subfigure}
     \begin{subfigure}[b]{0.22\linewidth}
        \includegraphics[width=\linewidth]{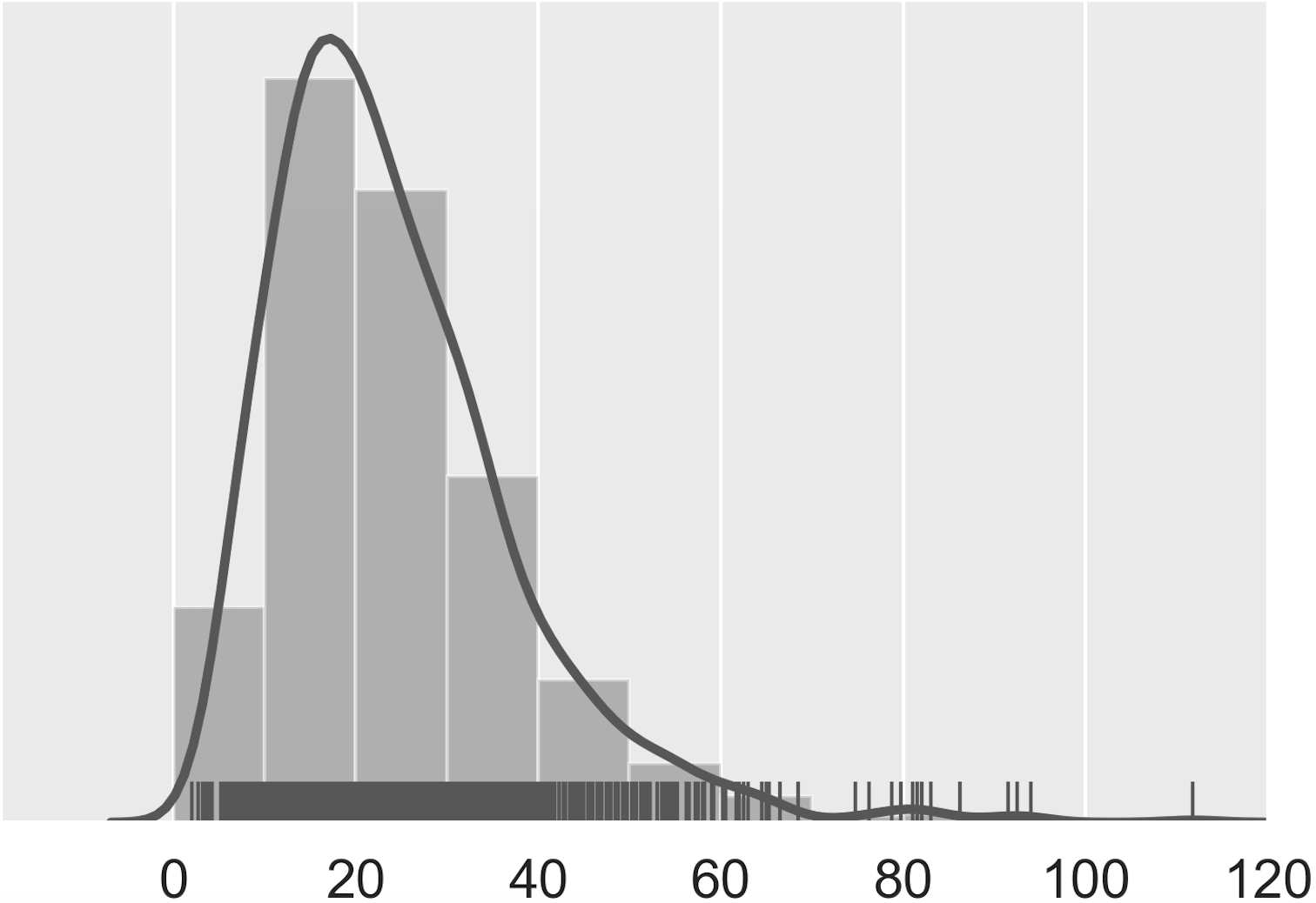}
        \caption{\english}
    \end{subfigure}
  \caption{Distribution of character-based Levenshtein distance between informal sentences and formal rewrites.}
  \label{fig:character_levem}
\end{figure*}
\begin{figure*}[ht!]
  \centering
  \begin{subfigure}[b]{0.24\linewidth}
    \includegraphics[width=\linewidth]{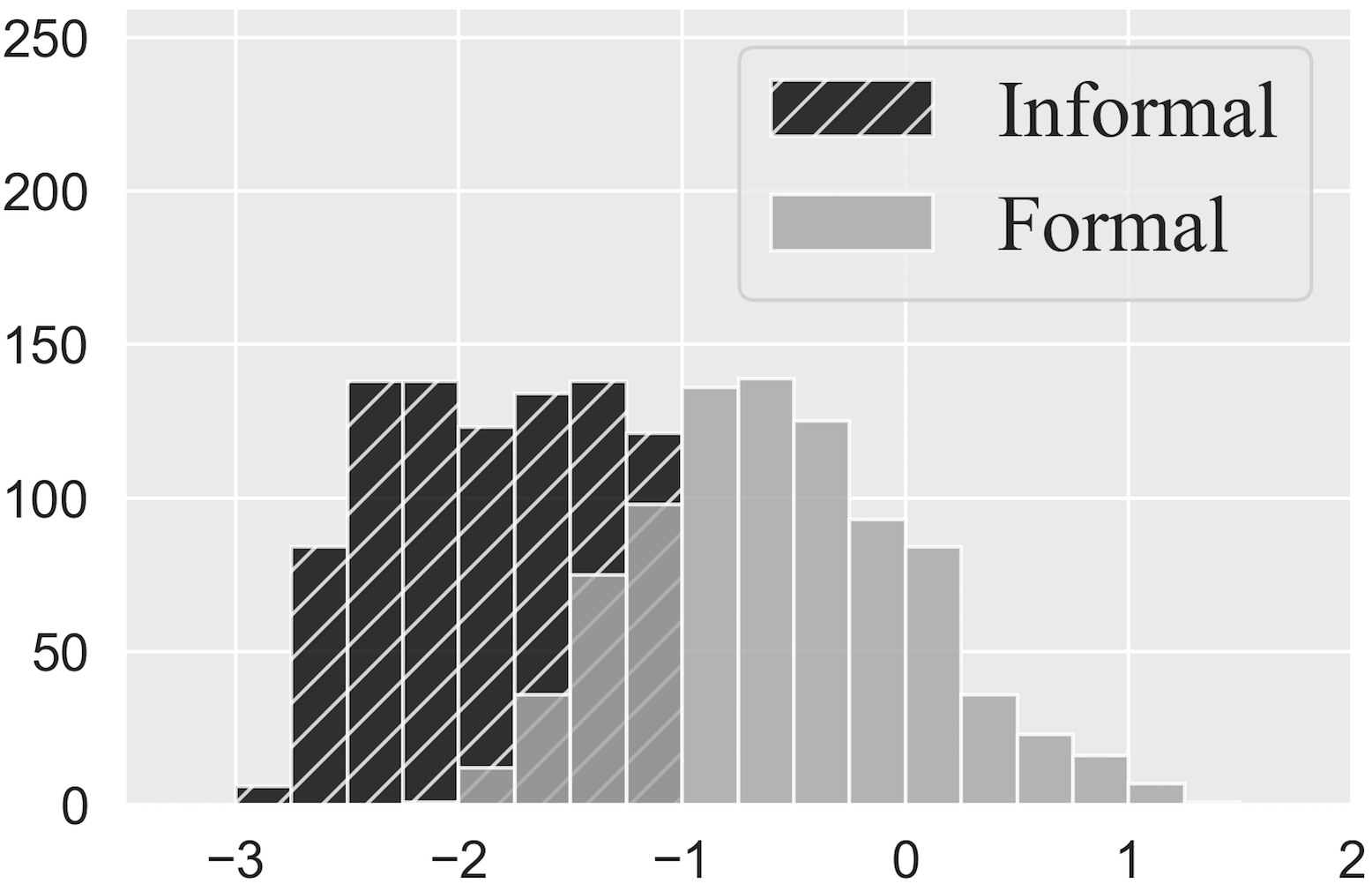}
    \caption{\portuguese}
  \end{subfigure}
    \begin{subfigure}[b]{0.22\linewidth}
        \includegraphics[width=\linewidth]{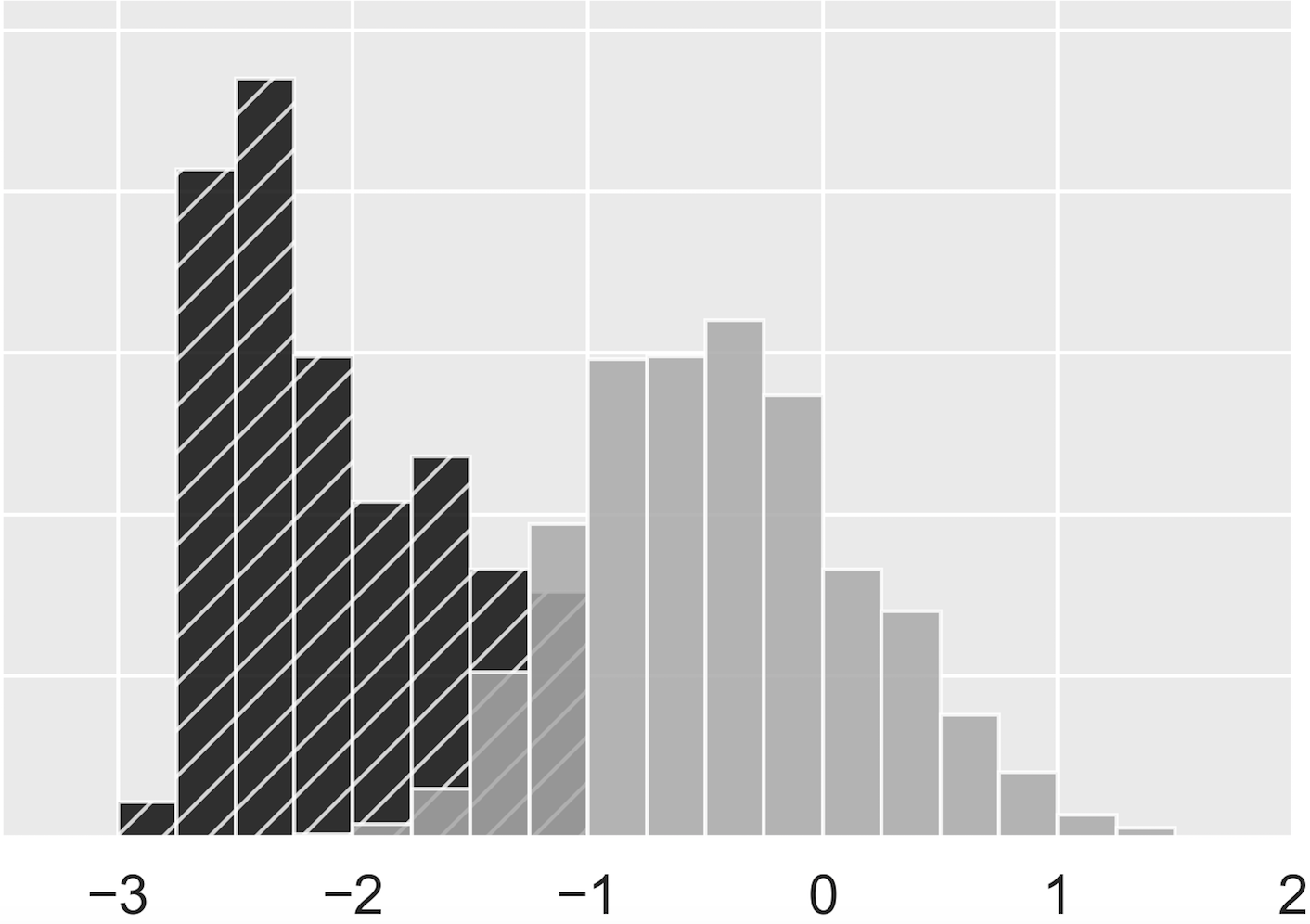}
        \caption{\french}
    \end{subfigure}
    \begin{subfigure}[b]{0.22\linewidth}
        \includegraphics[width=\linewidth]{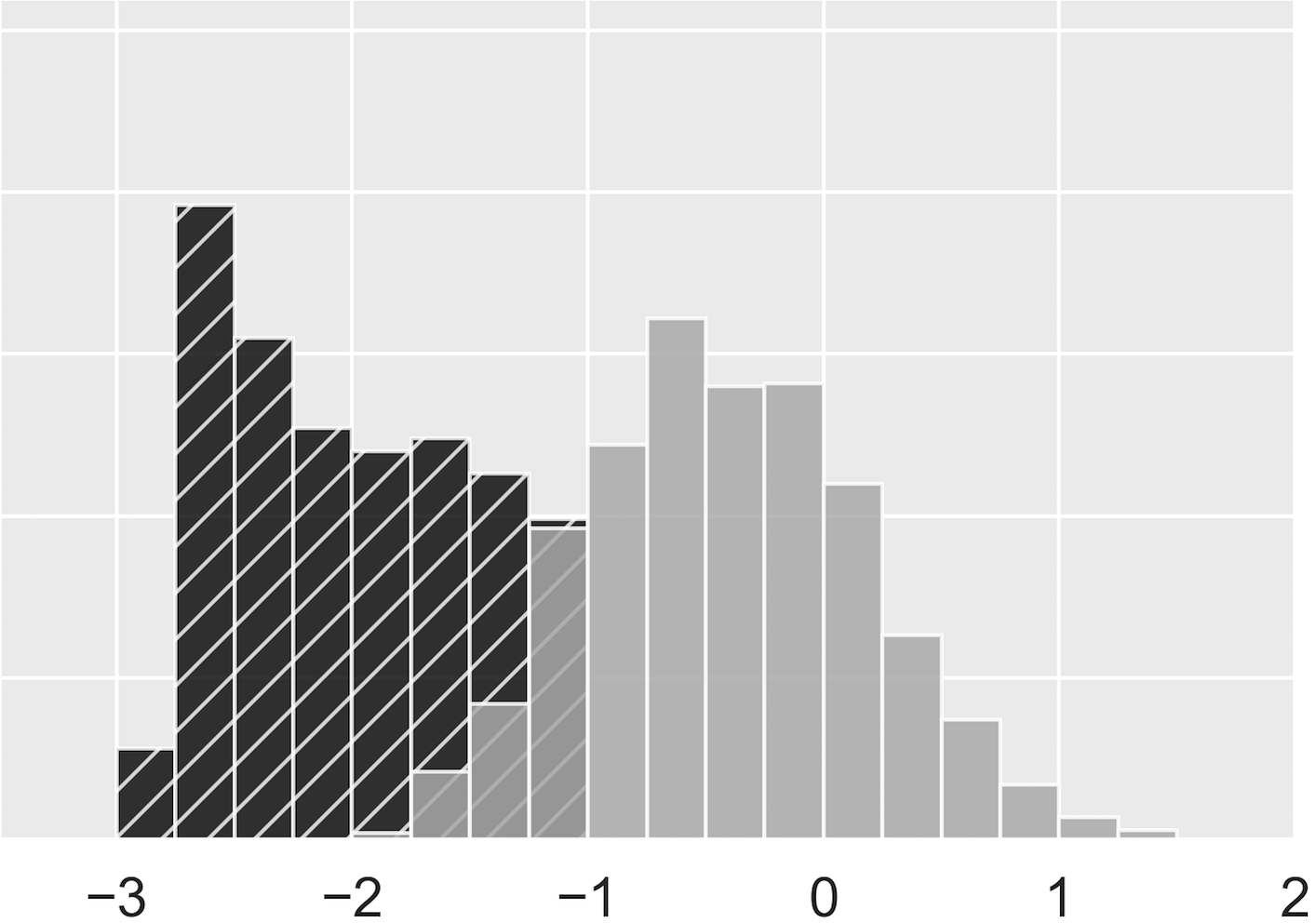}
        \caption{\italian}
    \end{subfigure}
    \begin{subfigure}[b]{0.22\linewidth}
        \includegraphics[width=\linewidth]{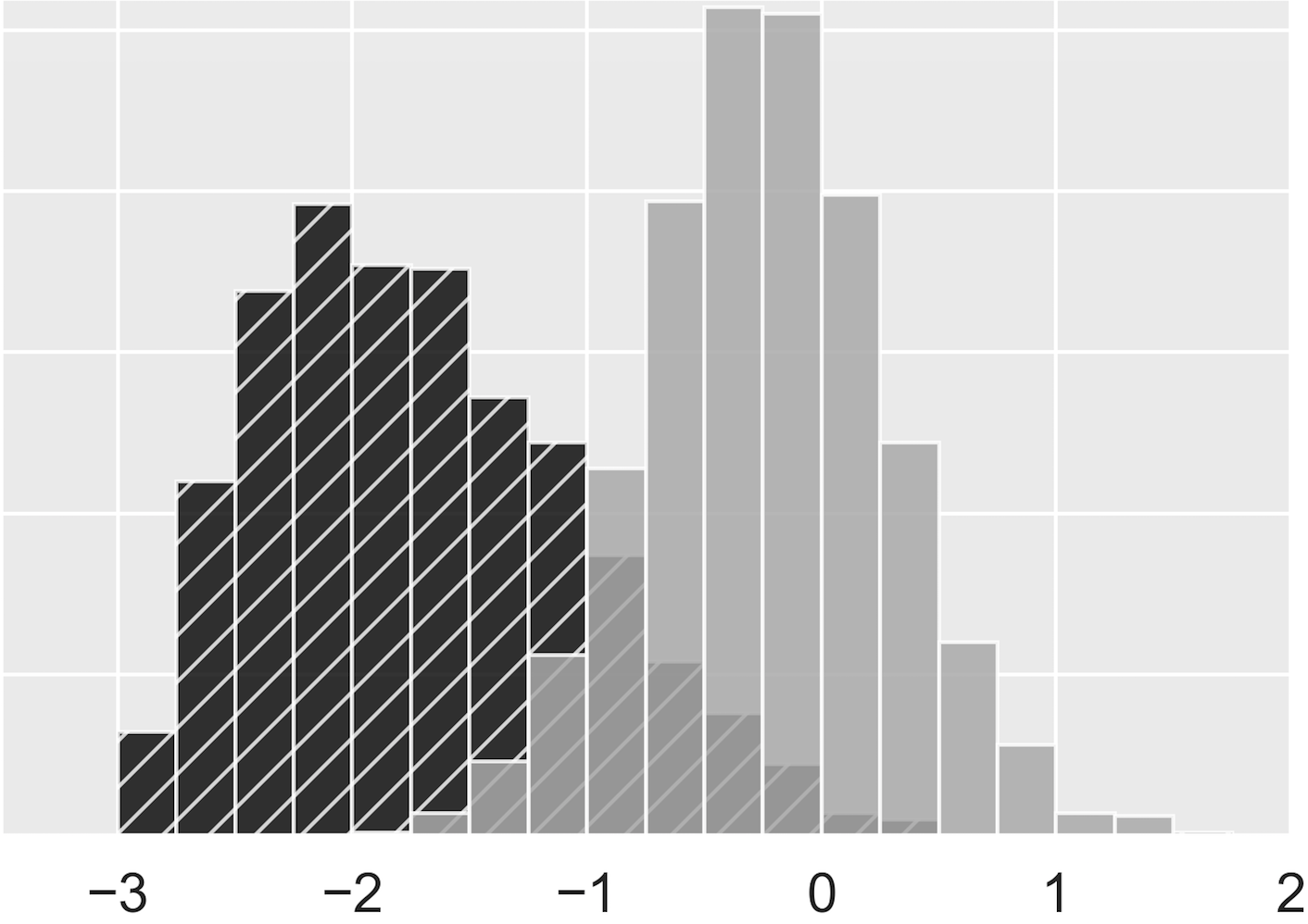}
         \caption{\english}
    \end{subfigure}
  \caption{Number of informal sentences and formal rewrites binned according to formality score.}
  \label{fig:shift}
\end{figure*}
\paragraph{Types of formality edits} Following \citet{pavlick-tetreault-2016-empirical}, we analyze the most frequent \textbf{edit} operations Turkers perform when formalizing the informal sentences. We conduct both an \textbf{automatic analysis} (details in \appendixmacro\ref{sec:quantitative_analysis_details}) of the whole set of rewrites, and a \textbf{human analysis} (details in \appendixmacro\ref{sec:qualitative_analysis_details}) of a random sample of $200$ rewrites per language (we recruited a native speaker for each language). 
Table~\ref{tab:types_of_edits} presents both analyses' results, where we also include the corresponding statistics for the English language (\gyafc). In general, we observe similar trends across languages: humans make edits covering both the "noisy-text" sense of formality (e.g., fixing punctuation, spelling errors, capitalization) and the more situational sense (paraphrase-based edits). Although cross-language trends are similar, we also observe differences: \textit{deleting fillers} and word \textit{completion} seems to be more prominent in the English rewrites than in other languages; \textit{normalizing} abbreviations is a considerably frequent edit type for Brazilian Portuguese; \textit{paraphrasing} is more frequent in the three non-English languages.
\begin{table}[ht!]
    \centering
    \scalebox{0.94}{
    
    \begin{tabular}{lrrrr}
    
    \rowcolor{gray!10}

    \textbf{\textsc{edit types}} & \textsc{br-pt} & \textsc{fr} & \textsc{it} &  \textsc{en} \\ 
    
    \toprule[1.2pt]
    \addlinespace[0.5em]

    \multirow{2}{*}{\textsc{capitalization}}         & $0.24$ & $0.33$ & $0.32$ & $0.43$ \\
                                                     & \C$0.43$ & \C$0.56$ & \C$0.60$ & \C$0.46$ \\ \addlinespace[0.15cm]
    \multirow{2}{*}{\textsc{punctuation}}            & $0.64$ & $0.87$ & $0.79$ & $0.74$ \\
                                                     & \C$0.76$ & \C$0.64$ & \C$0.66$ & \C$0.40$ \\ \addlinespace[0.15cm]
    \multirow{2}{*}{\textsc{spelling}}               & $0.36$ & $0.38$ & $0.32$ & $0.29$ \\
                                                     & \C$0.09$ & \C$0.12$ & \C$0.10$ & \C$0.14$ \\ \addlinespace[0.15cm]
    \multirow{2}{*}{\textsc{normalization}}          & $0.17$ & $0.01$ & $0.04$ & $0.07$  \\
                                                     & \C$0.27$ & \C$0.02$ & \C$0.06$ & \C$0.10$ \\ \addlinespace[0.15cm]
    \multirow{2}{*}{\textsc{split sentences}}        & $0.07$ & $0.06$ & $0.11$ & $0.08$ \\
                                                     & \C$0.07$ & \C$0.07$ & \C$0.07$ & \C$0.04$ \\ \addlinespace[0.15cm]
    \multirow{2}{*}{\textsc{paraphrase}}             & $0.68$ & $0.75$ & $0.86$ & $0.68$ \\
                                                     & \C$0.74$ & \C$0.75$ & \C$0.58$ & \C$0.47$ \\ \addlinespace[0.15cm]
    \textsc{unchanged}                               & $0.00$ & $0.00$ & $0.00$ & $0.01$  \\ 
    \textsc{lowercase}                               & $0.13$ & $0.06$ & $0.07$ & $0.13$ \\ 
    \textsc{repetition}                              & $0.01$ & $0.01$ & $0.02$ & $0.09$  \\ \addlinespace[0.15cm]
    \textsc{delete fillers}                          & \C$0.02$ & \C$0.05$ & \C$0.11$ & \C$0.26$ \\ 
    \textsc{completion}                              & \C$0.07$ & \C$0.04$ & \C$0.04$ & \C$0.15$  \\ 
    \textsc{add context}                             & \C$0.02$  & \C$0.00$  & \C$0.07$  & \cellcolor{gray!90}  \\ 
    \textsc{contractions}                            & \C$0.02$  & \C$0.01$  & \C$0.01$  & \cellcolor{gray!90}  \\ 
    \textsc{slang/idioms}                            & \C$0.19$ & \C$0.10$ & \C$0.15$    & \cellcolor{gray!90}    \\ 
    \textsc{politeness}                              & \C$0.82$  & \C$0.00$ & \C$0.10$   &  \cellcolor{gray!90}    \\ 
    \textsc{relativizers}                            & \C$0.01$   & \C$0.00$ & \C$0.00$  &  \cellcolor{gray!90}    \\ 
    \end{tabular}}
    
    \caption{Percentage categories of frequent edits calculated \textbf{automatically} (not highlighted) and \textbf{manually} (highlighted). Categories are not mutually exclusive.}
    \label{tab:types_of_edits}\vspace{-4mm}
\end{table}
\begin{table}[ht!]
    \centering
    \scalebox{0.94}{
    
    \begin{tabular}{lrrrr}
    
    \rowcolor{gray!10}

    \textbf{\textsc{metric}} & \textsc{br-pt} & \textsc{fr} & \textsc{it} &  \textsc{en} \\ 
    
    \toprule[1.2pt]
    \addlinespace[0.5em]

    LeD                                     & $0.44$  & $0.39$ & $0.54$ & $0.36$ \\
    \addlinespace[0.1em]
    \hline
    \addlinespace[0.3em]
    self-\bleu                             & $0.43$ & $0.35$ & $0.21$  & $0.51$ \\ 
    \end{tabular}}
    \caption{Surface differences \textbf{between} informal-formal pairs (LeD) and \textbf{within} formal rewrites (self-\bleu).}
    \label{tab:quantitative_diversity}\vspace{-4mm}
\end{table}
\paragraph{Surface differences of informal-formal pairs} We quantify surface-level differences between the informal sentences and formal rewrites via computing their character-level Levenshtein distance (Figure \ref{fig:character_levem}) and their pairwise Lexical Difference (LeD) based on the percentages of \textbf{tokens} that are 
\textit{not} found in both sentences~(Table~\ref{tab:quantitative_diversity}). Both analyses show that Italian rewrites have the most edits compared to their corresponding informal sentences. French and Brazilian Portuguese follow, with English rewrites being closer to the informal inputs.
\paragraph{Diversity of formal rewrites} Are Turkers making similar choices when formalizing text? Since a large number of reformulations consist of paraphrase-based edits (more than $50$\%), we want to quantify the extent to which the formal rewrites of each sentence are \textbf{diverse}, in terms of their lexical choices. To that end, we quantify diversity via measuring self-\bleu~\cite{zhu-2018}: considering one set of formal sentences as the hypothesis set and the others as references, we compute \bleu for each formal set and define the average \bleu score as a measure of the dataset's diversity. Higher scores imply less diversity of the set. Results (last row of Table~\ref{tab:types_of_edits}) show that \datasetname consists of more diverse rewrites compared to \gyafc.
\paragraph{Formality shift of rewrites} We analyze the formality distribution of the original informal sentences with their formal rewrites in \gyafc and \datasetname, as predicted by formality m\bert models (\S\ref{sec:automatic_eval}). The distributions of formal rewrites are skewed towards positive values~(Figure~\ref{fig:shift}).

\section{Multilingual \fst Experiments}
We benchmark eight \st models on  \datasetname to serve as baseline scores for future research. 
We describe the models~(\S\ref{sec:models}), the experimental setting~(\S\ref{sec:setting}), the human and automatic evaluation methods~(\S\ref{sec:automatic_eval} and \S\ref{sec:human_eval}), and results~(\S\ref{sec:results}). 
\subsection{Models}\label{sec:models}
\paragraph{Simple baselines} We define three baselines: \begin{inparaenum}
\item \textbf{\textsc{copy}} Motivated by ~\citet{pang-gimpel-2019-unsupervised} who notice that untransferred sentences with no alterations have the highest \bleu score by a large margin for \st tasks, we use this simple baseline as a lower bound;
\item \textbf{\textsc{rule-based}} Based on the quantitative analysis of \S\ref{sec:data_analysis} and similarly to~\sudhajoel, we 
develop a rule-based approach that
performs a set of predefined edits-based operations defined by hand-crafted rules. Example transformations include fix casing, remove repeated punctuation, handcraft a list of contraction expansions---a detailed description is found at \appendixmacro\ref{sec:rules_details};
\item \textbf{\textsc{round-trip mt}} Inspired by \citet{zhang-etal-2020-parallel} who identify useful training pairs from the paraphrases generated by round-trip translations of millions of sentences, we devise a simpler baseline that starts from a text in language $x$, pivots to \english and then backtranslates to $x$, using the \textsc{aws} translation service.\footnote{\url{https://aws.amazon.com/translate/}} 
\end{inparaenum}
\paragraph{NMT-based models with synthetic parallel data} We follow the \textsc{translate train}~\cite{conneau-etal-2018-xnli,artetxe-etal-2020-cross}
approach to collect data in multilingual settings:   
we obtain \textit{pseudo-parallel} corpora in each language via machine translating an \english resource of informal-formal pairs~(\S\ref{sec:setting}).\footnote{Details on the \textsc{aws} performance are found in \appendixmacro\ref{sec:aws_details}.}  Then, starting with \textsc{translate train} we benchmark the following \textsc{nmt}-based models:
\begin{inparaenum}
\item \textbf{\textsc{translate train tag}} extends a leading \english \fst approach \cite{niu-etal-2018-multi}
and trains a unified model that handles either formality direction via attaching a source tag that denotes the desired target formality;
\item \textbf{\textsc{multi-task tag-style}}  ~\citet{niu-etal-2018-multi} augments the previous approach with bilingual data that is automatically identified as formal (\S\ref{sec:automatic_eval}).   The models are then trained in a multi-task fashion;
\item \textbf{\textsc{backtranslate}} augments the \textsc{translate train} data with
back-translated sentences of automatically detected informal text~\cite{sennrich-etal-2016-improving}, using $1.$ as the base model. We exclude backtranslated pairs consisting of copies. \end{inparaenum} The output of the \textsc{rule-based} system is given as input to each model at inference time. For all three models, we run each system with $4$ random seeds, and combine them in a linear ensemble for decoding.
\paragraph{Unsupervised approaches} 
We benchmark two unsupervised methods that are used for \english \st:
\begin{inparaenum}
\item \textbf{\textsc{unpsupervised neural machine translation (unmt)}} \cite{lample-2019-multi} defines a pseudo-supervised setting and combines denoising auto-encoding and back-translation losses;
\item \textbf{\textsc{deep latent sequence model (dlsm)}} \cite{he2020a} defines a probabilistic generative story that treats two unpaired corpora of separate styles as a partially observed parallel corpus and learns a mapping between them, using variational inference.
\end{inparaenum}  
\subsection{Experimental setting}\label{sec:setting}
\paragraph{Training data} For \textsc{translate train tag} we use \textsc{gyafc}, a large set of $110$K  \english informal-formal parallel sentence-pairs obtained through crowdsourcing. 
Additionally, we augment the translated resource with OpenSubtitles ~\cite{Lison2016OpenSubtitles2016EL} bilingual data used for training \textsc{mt} models.\footnote{Data are available at: \url{http://opus.nlpl.eu/}.} 
Given that bilingual sentence-pairs can be noisy, we perform a filtering step to extract noisy bitexts using the Bicleaner toolkit~\cite{prompsit:2018:WMT}.\footnote{We use the publicly available pretrained Bicleaner models: \url{https://github.com/bitextor/bicleaner}, and discard sentences with a score lower than $0.5$.} Furthermore, we apply the same filtering steps as in \S\ref{sec:data_collection}~(Curation rational). Finally, each of the remaining sentences is assigned a formality score~(\S\ref{sec:automatic_eval}), resulting in two pools of informal and formal text. Training instances are then randomly sampled from those pools: formal parallel pairs are used for \textsc{multi-task tag-style}; 
informal target side sentences are backtranslated for \textsc{backtranslate}; both informal and formal target-side texts are independently sampled from the two pools for training unsupervised models. Finally, for unsupervised \fst in \french, we additionally experiment with in-domain data from the L$26$ French Yahoo! Answer Corpus that consists of $1.7$M \french questions. \footnote{\url{https://webscope.sandbox.yahoo.com/catalog.php?datatype=l&did=74}}\textsuperscript{,}
\footnote{Split into $6.2$M/$6.6$M formal/informal sentences.} 
Table~\ref{tab:training_data_sizes} includes statistics on training sizes.\footnote{Bilingual data statistics are in ~\appendixmacro\ref{sec:opensubs_extra_stats}.} 
\begin{table}[h]
    \centering
    \scalebox{0.74}{
    \begin{tabular}{lccc}

    \rowcolor{gray!10}
    \textbf{\textsc{methods}} & \textsc{gyafc} & OpenSubs. & L6 Yahoo! \\   
     \toprule[1.2pt]
    \addlinespace[0.5em] 

    \textsc{transl. train tag}       &   $110$K  &   $-$     &  $-$  \\
    \textsc{multi-task tag-style}    &   $110$K  & \hspace{0.6em}  $2$M &  $-$  \\    
    \textsc{backtranslate}   &   $110$K  & \hspace{0.6em}  $2$M &  $-$  \\
    
    \addlinespace[0.5em] 
    
    \textsc{unmt/dlsm}               &    $-$    &  $110$K   &  $-$  \\  
    \textsc{unmt/dlsm (in-domain)}   &    $-$    &   $-$     &  $2$M  \\  
    \end{tabular}}
    \caption{Number of training instances for each model.}\vspace{-4mm}
    \label{tab:training_data_sizes}
\end{table}
\paragraph{Preprocessing} We preprocess data consistently across languages using \textsc{moses}~\cite{koehn-etal-2007-moses}. Our pipeline consists of three steps: a) normalization; b) tokenization; c) true-casing. For \textsc{nmt}-based approaches, we also learn joint source-target \textsc{bpe} with $32$K operations \cite{sennrich-etal-2016-improving}.
\paragraph{Model Implementations} For NMT-based and unsupervised models we use the open-sourced impementations of \citet{niu-etal-2018-multi} and \citet{he2020a}, respectively.\footnote{\url{https://github.com/xingniu/multitask-ft-fsmt}}\textsuperscript{,}\footnote{ \url{https://github.com/cindyxinyiwang/deep-latent-sequence-model}}
We include more details on model architectures in \appendixmacro\ref{sec:nmt_details}.
\begin{table*}[!ht]
    \centering
    \scalebox{0.6}{
    \begin{tabular}{cl@{\hskip 0.5in}lll@{\hskip 0.5in}lll@{\hskip 0.5in}lll@{\hskip 0.5in}lll}

    & & \multicolumn{3}{c}{self-\textsc{bleu} ($\shneg$)} & \multicolumn{3}{c}{\textsc{mbert score} ($\shneg$)} & \multicolumn{3}{c}{\textsc{perplexity} ($\shpos$)} & \multicolumn{3}{c}{multi-\textsc{bleu} ($\shneg$)}\\ 
    
    & \textbf{Method}         & \portuguese & \italian   & \french   & \portuguese & \italian & \french   & \portuguese & \italian & \french  & \portuguese & \italian & \french\\\
    & \textsc{gold standard}  & \ptsix$23.8$ & \itseven$16.2$  & \freight$19.2$     & \ptzero$-0.48$ & \itzero$-0.39$ & \frzero$-0.42$ & \ptzero$5.32$ & \itone$5.84$ & \frzero$4.70$ & \fil      & \fil        & \fil \\
    & \textsc{copy}           & \fil         & \fil        & \fil      & \ptsix$-1.67$   & \iteight$-1.94$  & \frsix$-2.04$  & \pteight$8.77$   & \iteight$8.58$  & \frsix$8.36$  & \ptfive$46.2$  & \itfive$40.2$ & \frfive$44.3$\\
    & \textbf{\textsc{rule-based}}         & \ptone$\mathbf{79.8}$ & \itone$\mathbf{83.4}$ & \frone$\mathbf{85.4}$ &  \ptfive$-1.33$  & \itfive$-1.41$ & \frfive$-1.59$ & \ptseven$7.21$  & \itsix$7.24$ & \frfive$7.07$ &  \ptthree$53.1$ & \itone$\mathbf{43.1}$  & \frthree$46.1$   \\
    & \textbf{\textsc{round-trip mt}}           & \ptfive$33.0\SG$ & \itfive$39.7\SG$ & \frfive$32.5\SG$ & \ptthree$-1.10\SG$ & \itfour$-1.27\SG$ & \frfour$-1.20\SG$ & \ptthree$6.27\SG$ & \itfour$6.84\SG$ & \frfour$6.03\SG$ & \ptsix$43.0\SG$   & \itsix$34.5\SG$  & \frsix$41.4\SG$ \\  
    & \textbf{\textsc{translate train tag}}          & \ptthree$60.9\SG$& \itthree$57.0\SG$ & \frthree$51.8\SG$  & \pttwo$-0.96\SG$  & \itone$\mathbf{-0.90}\SG$   & \frtwo$-0.94\SG$   & \pttwo$5.89\SG$  & \ittwo$6.03\SG$   & \frtwo$5.36\SG$ & \ptfour$52.4\sg$  & \itthree$42.6$  & \frfour$44.9\sg$ \\
    \rot{\rlap{~Automatic}}
    & \textbf{\textsc{multi-task tag-style}}   & \ptfour$57.6\SG$ & \itfour$54.6\SG$  & \frfour$48.8\SG$ & \ptone$\mathbf{-0.91}\SG$  & \ittwo$-0.91\SG$   & \frone$\mathbf{-0.86}\SG$   & \ptone$\mathbf{5.85}\SG$  & \itzero$\mathbf{5.83}\SG$  & \frone$\mathbf{5.19}\SG$ & \ptone$\mathbf{55.0}\sg$   & \itfour$42.5$   & \frtwo$47.9\sg$ \\
    & \textbf{\textsc{backtranslate}}     & \pttwo$75.7\SG$  & \ittwo$68.8\SG$   & \frtwo$71.9\SG$   & \ptfour$-1.12\SG$ & \itthree$-1.07\SG$ & \frthree$-1.19\SG$ & \ptfour$6.39\SG$ & \itthree$6.28\SG$ & \frthree$5.93\SG$ & \pttwo$54.6\SG$   & \ittwo$42.9$    & \frone$\mathbf{48.3}\SG$\\
    
    & \textbf{\textsc{unmt}} & \pteight$15.3\SG$ & \iteight$15.1\SG$ & \frseven$19.7\SG$ & \ptseven$-1.74\SG$ & \iteight$-1.84\SG$ & \freight$-1.91\SG$ & \ptsix$6.42\SG$ & \itsix$6.40\SG$ & \frfive$6.61\SG$ & \pteight$14.8\SG$ & \iteight$11.5\SG$ & $17.0\SG$ \\
    & \textbf{\textsc{dlsm}} & \ptseven$20.4\SG$ & \itsix$19.6\SG$ & \frsix$21.3\SG$ & \ptseven$-1.84\SG$ & \itsix$-1.61\SG$ & \freight$-1.90\SG$ & \ptseven$6.97\SG$ & \itseven$7.45\SG$ & \frfive$6.97\SG$ & \ptseven$18.4\SG$ & \itseven$12.4\SG$ & \freight$17.4\SG$ \\

    & \textbf{\textsc{unmt (in-domain)}} & \fil  & \fil  & \frthree$52.8\SG$ & \fil  & \fil  & \frseven$-1.71\SG$ & \fil & \fil & \frfive$5.69\SG$ & \fil  & \fil & \frseven$35.0\SG$ \\
    & \textbf{\textsc{dlsm (in-domain)}} & \fil  & \fil  & \frtwo$71.2\SG$ & \fil & \fil & \frsix$-1.66$ & \fil & \fil  & \frsix$6.78\SG$ & \fil  & \fil  & \frsix$42.1\SG$ \\  

    \addlinespace[3.5em]

    &    & \multicolumn{3}{l}{\textsc{mean. preserv.} ($\shneg$)} & \multicolumn{3}{c}{\textsc{formality} ($\shneg$)} & \multicolumn{3}{c}{\textsc{fluency} ($\shneg$)} & \multicolumn{3}{c}{\textsc{overall} ($\shneg$)} \\ 
    &         & \portuguese      & \italian        & \french        & \portuguese     & \italian         & \french         & \portuguese & \italian & \french & \portuguese & \italian & \french\\\
    & \textsc{gold standard}  & \ptfive$4.88$    & \itfive$5.18$   & \frthree$5.08$ & \ptzero$+0.66$   & \itzero$+1.46$    & \frzero$+1.14$   & \ptzero$4.54$ & \itzero$4.79$  & \frzero$4.51$ & \fil & \fil & \fil\\
    & \textsc{copy}           & \fil             & \fil            & \fil           & \ptsix$-1.18$   &  \itsix$-1.06$   & \frsix$-0.75$   & \ptsix$3.91$  & \itfive$4.09$  & \frfive$3.88$  & \fil & \fil & \fil \\
    & \textbf{\textsc{rule-based}}     & \ptone$\mathbf{5.70}$     & \itthree$\mathbf{5.96}$  & \frone$\mathbf{5.95}$   & \ptfive$-0.51$  &  \itfive$-0.36$  & \frfive$-0.36$  & \ptfive$4.06$ & \itthree$4.24$ & \frfour$3.94$     & \ptfour$2.81$ & \itone$3.23$ & \frthree$2.89$\\
    & \textbf{\textsc{round-trip mt}}       & \ptsix$4.80\SG$     & \itsix$5.03\SG$   & \frsix$4.50\SG$    & \ptone$\mathbf{-0.17}\SG$   &  \itfour$-0.34$  & \frone$\mathbf{+0.25}\SG$      & \ptfour$4.07$ & \itsix$4.03\sg$   & \frthree$3.97$ & \ptthree$2.89$ & \itfour$2.54\SG$ & \frtwo$2.94$\\
          \rot{\rlap{~Human}}
    & \textbf{\textsc{translate train tag}}        & \ptfour$4.97\SG$   & \itfive$5.18\SG$   & \frfive$4.56\SG$  & \pttwo$-0.26\sg$   & \itone$\mathbf{+0.07}\sg$      & \frtwo$+0.10\SG$  & \ptthree$4.15$ & \ittwo$4.33$  & \frsix$3.84$   & \ptone$3.03\sg$ & \ittwo$3.02\sg$ & \frfour$2.83$\\
    & \textbf{\textsc{multi task tag-style}} &  \ptthree$5.07\SG$ & \itfive$5.18\SG$   & \frfour$4.81\SG$  & \ptthree$-0.27$ &  \ittwo$+0.01\sg$     & \frthree$+0.05\SG$   & \ptthree$4.15$ & \itfour$4.22$ & \frfive$3.88$  & \ptzero$\mathbf{3.24}\SG$ & \itthree$2.93\sg$ & \frzero$\mathbf{3.19}\sg$ \\
    & \textbf{\textsc{backtranslate}}   &  \pttwo$5.54\SG$   & \itfour$5.72\SG$   & \frtwo$5.59\SG$   & \ptfour$-0.49$  &  \itthree$-0.16$  & \frfour$-0.13$         & \pttwo$\mathbf{4.19}\sg$   & \itone$\mathbf{4.37}$  & \frtwo$\mathbf{4.00}$   & \pttwo$3.01\sg$ & \itzero$\mathbf{3.25}$ & \frone$3.13\SG$\\
    
    \end{tabular}}
    \caption{Automatic and human evaluation results for multilingual \fst. 
    $\sg$ denotes statistical significance differences compared to \textsc{rule-based} ($p<0.05$).
    \textsc{nmt}-based results are ensembles of $4$ systems; unsupervised models are average results across $4$ reruns. Darker colors denote higher rankings, bold numbers denote best systems.}
    \label{tab:results_all}
\end{table*}

\ignore{
\begin{figure*}[ht!]
  \centering
  \begin{subfigure}[b]{0.26\linewidth}
    \includegraphics[width=\linewidth]{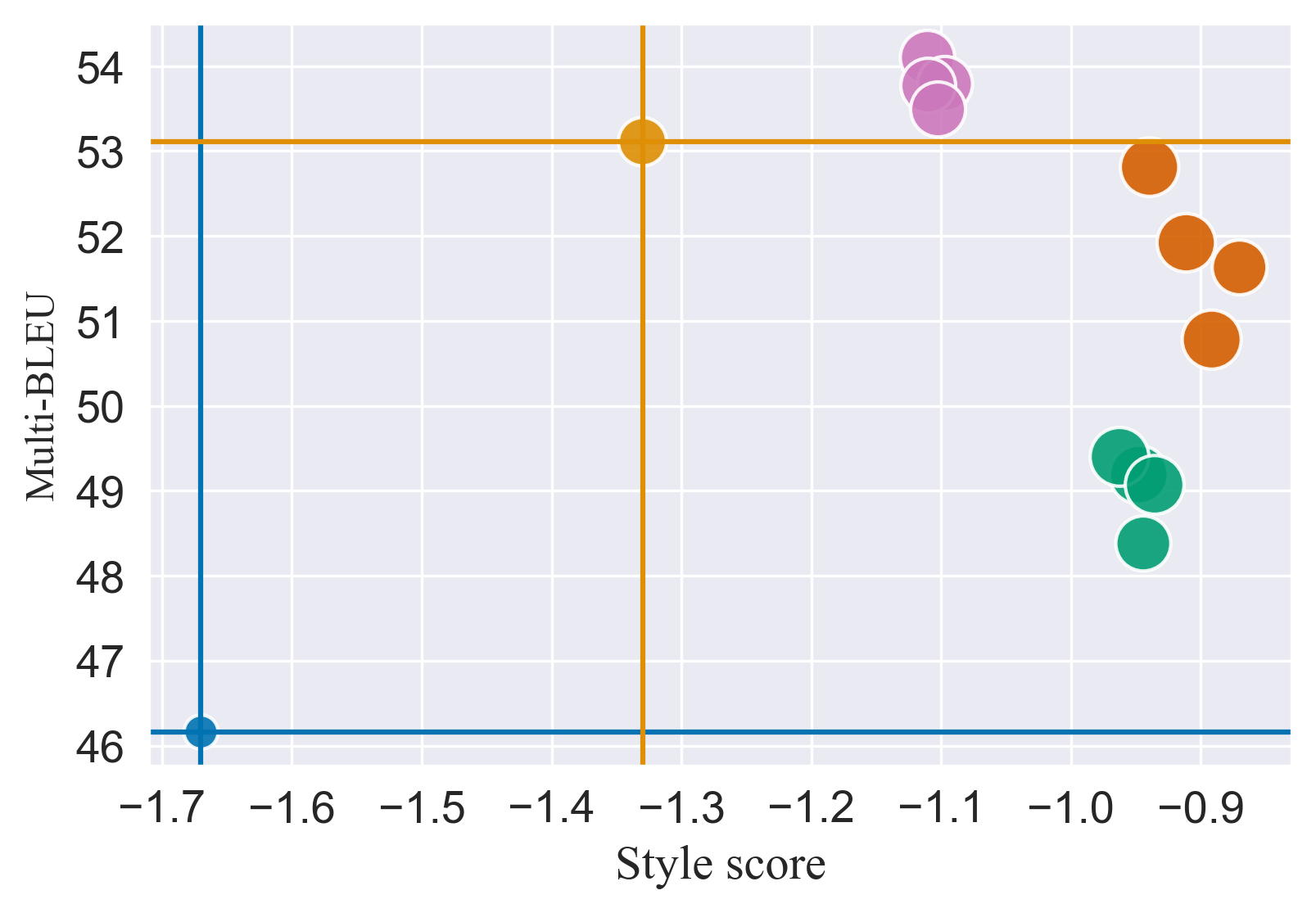}
     \caption{\portuguese}
  \end{subfigure}
  \begin{subfigure}[b]{0.25\linewidth}
    \includegraphics[width=\linewidth]{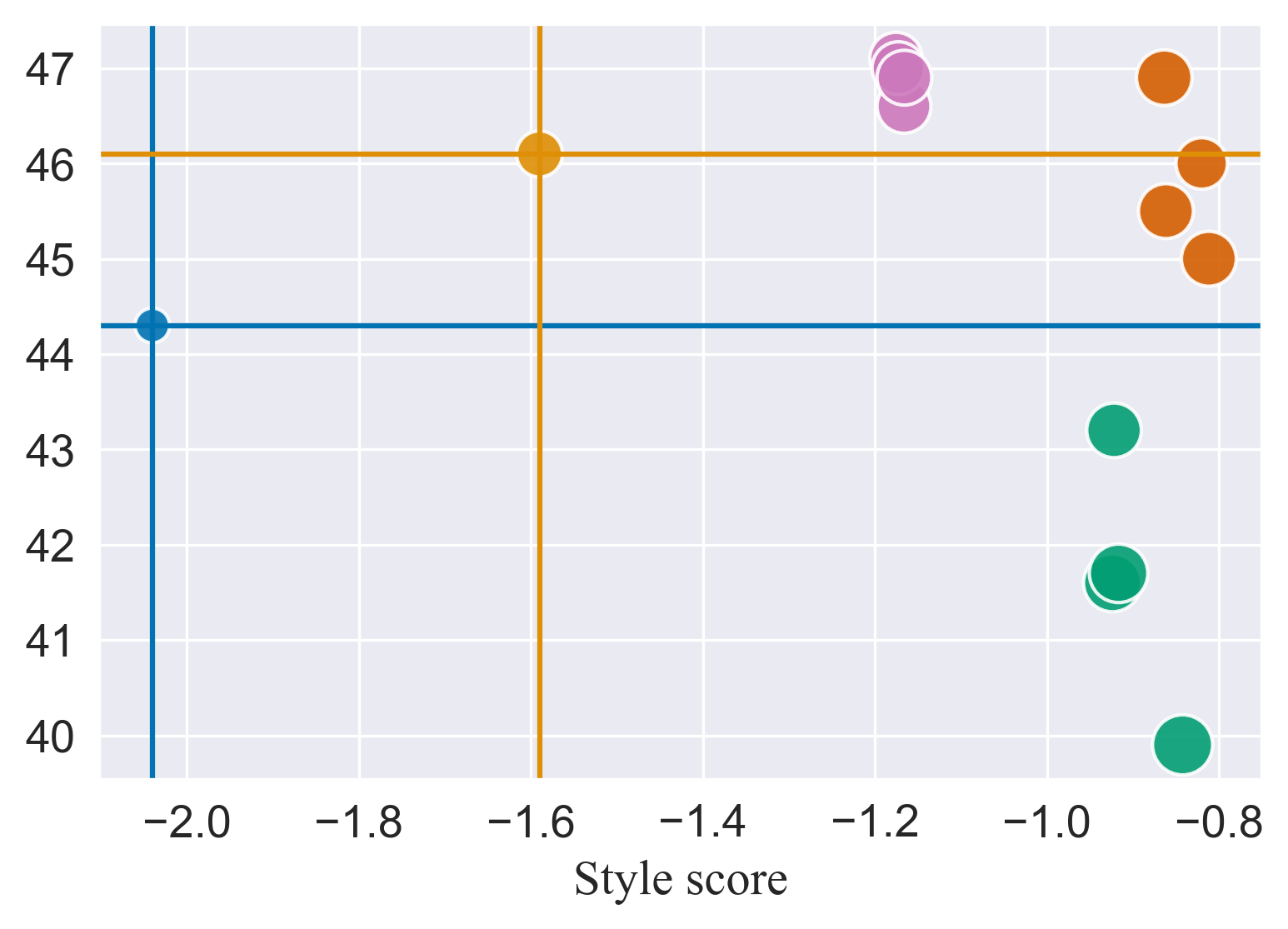}
    \caption{\french}
  \end{subfigure}
  \begin{subfigure}[b]{0.25\linewidth}
    \includegraphics[width=\linewidth]{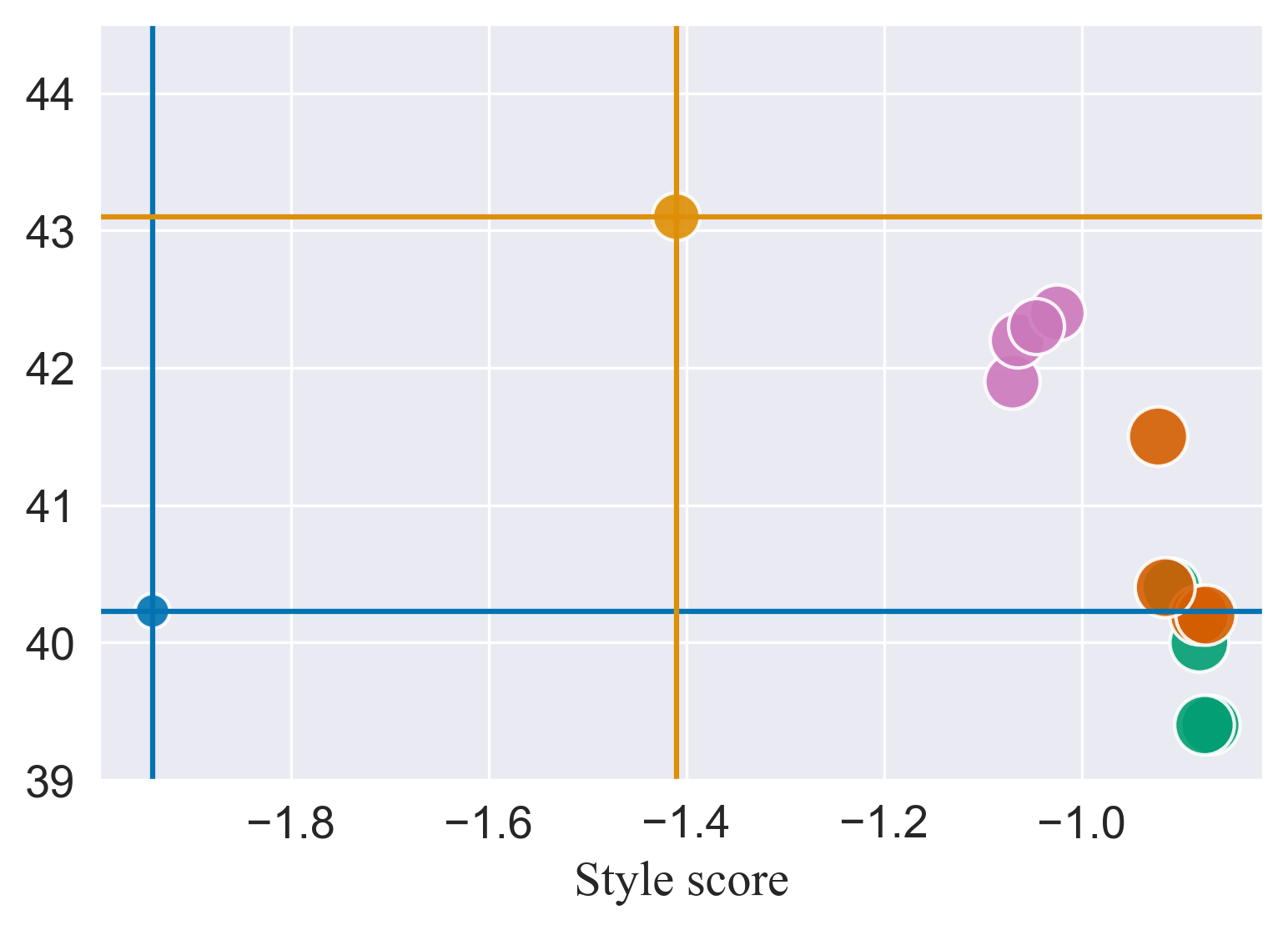}
    \caption{\italian}
  \end{subfigure}
    \begin{subfigure}[b]{0.12\linewidth}
    \includegraphics[width=\linewidth]{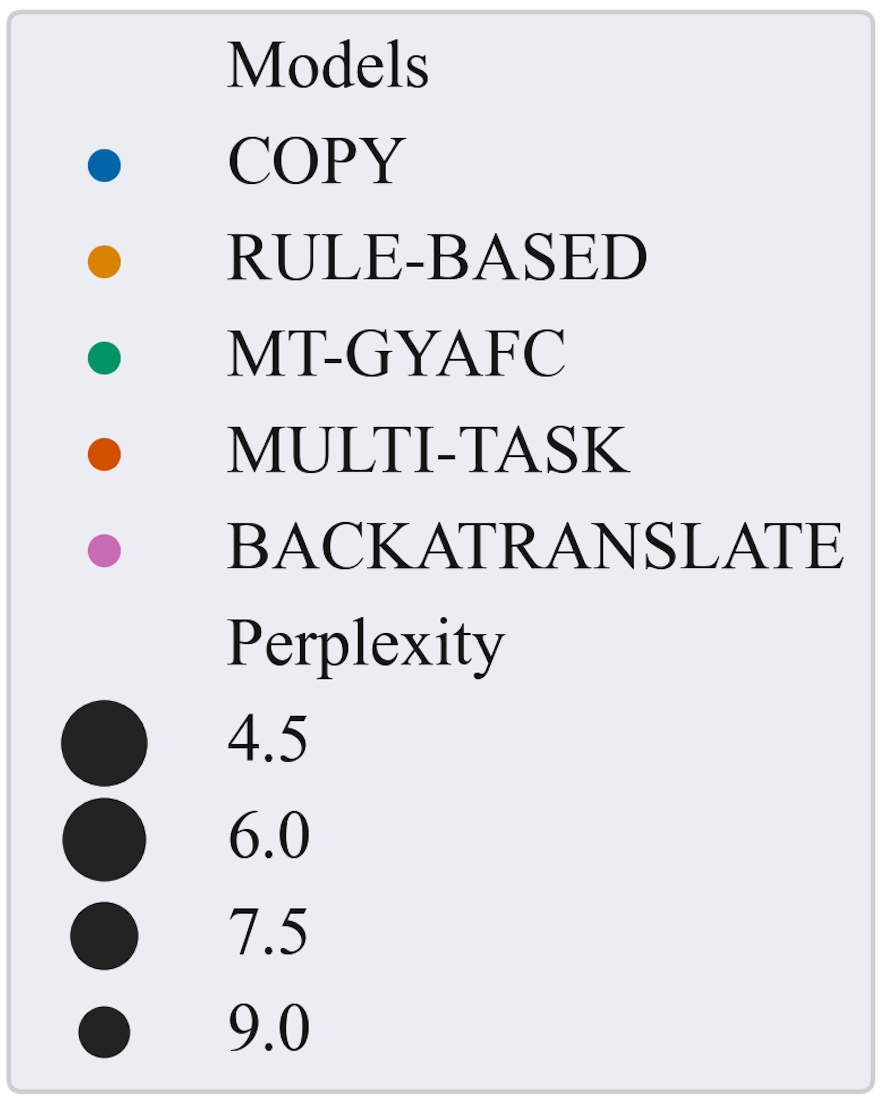}
    \vspace{1.3em}
  \end{subfigure}
  \caption{\textbf{\bleu vs.\ \textsc{style score} vs.\ \textsc{perplexity}} trade-off plots across $4$ reruns.}
  \label{fig:tradeoff_blue_vs_style_score}
\end{figure*}}
\subsection{Automatic Evaluation}\label{sec:automatic_eval}
\noindent
Recent work on \st evaluation highlights the \textbf{lack of standard evaluation practices}~\cite{Yamshchikov2020StyletransferAP,pang-2019-towards,pang-gimpel-2019-unsupervised, mir-etal-2019-evaluating}. We follow the most frequent evaluation metrics used in \english tasks and  measure the quality of the system's outputs with respect to four dimensions, while we leave an extensive evaluation of automatic metrics for future work.
\paragraph{Meaning Preservation} We compute self-\textsc{bleu}~\cite{bleu} which compares system outputs with the informal sentences. 
\paragraph{Formality} We average the style transfer score of transferred sentences computed by a formality regression model. We fine-tune m\bert~\cite{devlin-etal-2019-bert} pre-trained language models on the machine-translated answers genre from~\citet{pavlick-tetreault-2016-empirical} that consists of about $4$K human-annotated sentences rated on a $7$-point formality scale. To acquire an annotated corpus in the languages of interest, we follow the \textsc{translate train} transfer approach: we propagate the original \english training data's human ratings to their corresponding translations, assuming that translation preserves formality.\footnote{See~\appendixmacro\ref{sec:formality_artifacts} for discussion on this assumption.} To evaluate the multilingual formality regression models' performance, we crowdsourced human judgments of $5$ Turkers for $200$ sentences per language. We report Spearman correlations of $64$ (\portuguese), $70$ (\italian), $71$ (\french), and $81$ (\english).
\paragraph{Fluency} We compute the logarithm of each sentence's probability---computed by a $5$-gram Kneser-Ney language model~\cite{kneser-ney-1995}---and normalize it by the sequence length. We train each language model on $2$M random sample of the non-English side of OpenSubtitles formal data.
\paragraph{Overall} We compute multi-\textsc{bleu}~\cite{post-2018-call}
via comparing with multiple formal rewrites on \datasetname. 
\citet{freitag-etal-2020-bleu} shows that correlation with human judgments improves when considering multiple references for \textsc{mt} evaluation.
\subsection{Human evaluation}\label{sec:human_eval}
Given that automatic evaluation of \st lacks standard evaluation practices---even in cases when \english is considered---we turn to human evaluation to reliably assess our baselines following the protocols of~\sudhajoel. We sample a subset of $100$ sentences from \datasetname per language, evaluate outputs of $5$ systems, and collect $5$ judgments per instance.We open the task to all workers passing~\textbf{\textsc{qc2}} in Table~\ref{tab:worker_statistics}. 
We include inter-annotator agreement results in~\appendixmacro\ref{sec:human_agreement}.
\paragraph{Formality} We collect formality ratings for the original informal reference, the formal human rewrite, and the formal system outputs on a $7$-point discrete scale of $-3$ to $3$, following~\citet{Lahiri15} (\textit{Very informal}$\rightarrow$~\textit{Informal}$\rightarrow$~\textit{Somewhat Informal}$\rightarrow$~\textit{Neutral}$\rightarrow$~\textit{Somewhat Formal}$\rightarrow$~\textit{Formal}~$\rightarrow$~\textit{Very Formal}).
\paragraph{Fluency}  We collect fluency ratings for the original informal reference, the formal human rewrite, and the formal system outputs on a discrete scale of $1$ to $5$, following~\citet{heilman2014predicting}~
(\textit{Other}~$\rightarrow$~
\textit{Incomprehensible}~$\rightarrow$~
\textit{Somewhat Comprehensible}~$\rightarrow$~
\textit{Comprehensible}~$\rightarrow$~
\textit{Perfect}). 
\paragraph{Meaning Preservation} We adopt the annotation scheme of Semantic
Textual Similarity~\cite{agirre-etal-2016-semeval}: given the informal reference and formal human rewrite or the formal system outputs, Turkers rate the two sentences' similarity on a $1$ to $6$ scale
(\textit{Completely dissimilar}~$\rightarrow$~
\textit{Not equivalent but on same topic}~$\rightarrow$~
\textit{Not equivalent but share some details}~$\rightarrow$~
\textit{Roughly equivalent}~$\rightarrow$~
\textit{Mostly equivalent}~$\rightarrow$~
\textit{Completely equivalent}). 
\paragraph{Overall} We collect overall judgments of the system outputs using \textit{relative ranking}: given the informal reference and a formal human rewrite, workers are asked to rank system outputs in 
the order of their overall formality, taking into account both fluency and meaning preservation. An overall score is then computed for each model 
via averaging results across annotating instances. 
\subsection{Results}\label{sec:results}
Table~\ref{tab:results_all} shows automatic results for all models across the four dimensions as well as human ratings for selected top models.  
\paragraph{\textsc{nmt}-based model evalatuation} 
Concretely, the \textsc{rule-based} baselines are significantly ($p<0.05$) the best performing models in terms of \textbf{meaning preservation} across languages. This  result is intuitive as the \textsc{rule-based} models act at the surface level and are unlikely to change the informal sentence's meaning. The \textsc{backtranslate} ensemble systems are the second-best performing models in terms of meaning preservation, while the \textsc{round-trip mt} outputs diverge semantically from the informal sentences the most. Those results are consistent across languages and human/automatic evaluations. On the other hand, when we compare systems in terms of their \textbf{formality}, we observe the opposite pattern: the \textsc{rule-based} and \textsc{backtranslate} outputs are the most informal compared to the other ensemble \textsc{nmt}-based approaches across languages.  Interestingly,  the \textsc{round-trip mt} outputs exhibit the largest formality shift for \portuguese and \french as measured by human evaluation.   The trade-off between meaning preservation and formality among models was also observed in \english (\sudhajoel).
Moreover, when we move to \textbf{fluency}, we notice similar results across systems. Specifically, human evaluation assigns almost all models an average score of $>4$, denoting that system outputs are comprehensible on average, with small differences between systems not being statistically significant. Notably, perplexity tells a different story: all system outputs are significantly better compared to the  \textsc{rule-based} systems across configurations and languages. This result denotes that perplexity might not be a reliable metric to measure fluency in this setting, as noticed in \citet{mir-etal-2019-evaluating} and \citet{ krishna-etal-2020-reformulating}. When it comes to the \textbf{overall ranking} of systems, we observe that the \textsc{nmt}-based ensembles are better than the \textsc{rule-based} baselines for \portuguese and \french, yet by a small margin as denoted by both multi-\bleu and human evaluation. However, the corresponding results for \italian denote that there is no clear win, and the \textsc{nmt}-based ensembles still fail to surpass the naive \textsc{rule-based} models, yet by a small margin. Finally, all ensembles outperform the trivial \textsc{copy} baseline. Table~\ref{tab:system_outputs} presents examples of system outputs.
As a side note, we followed the recommendation of \citet{tikhonov-yamshchikov-2018-sounds} to show the performance of \st models of individual runs and visualize trade-offs between metrics better.  Unlike their work which found that reruns of the same model showed wide performance discrepancies, we found that most of our \textsc{nmt}-based models did not vary in performance on \datasetname.  The results can be visualized in \appendixmacro\ref{sec:trade_offs_}. 
\begin{table*}[!ht]
    \centering
    \scalebox{0.8}{
    \begin{tabular}{lll}
    
    \toprule[2pt]
    \rowcolor{gray!10}
    (\portuguese)
    & \small \textbf{\textsc{informal}}  &  n preciso pedir pois sei q ela vai vir atras!!                     \\ 
     \rowcolor{gray!10}
    & \small \textbf{\textsc{xformal rewrite}}   & Não é preciso pedir, pois sei que ela virá em busca! \\

    \addlinespace[2mm]
    & \small \textsc{rule-based}            &  {\dg {não}} preciso pedir pois sei {\dg {que}} ela vai vir atras{\dg {!}} \\
    & \small \textsc{round-trip mt}         &  {\dg {Não}} preciso {\dg {perguntar porque}} sei {\dg {que}} ela {\dg {virá atrás de mim!}}  \\
    & \small \textsc{translate train tag}   &  {\dg {Eu não}} preciso pedir {\dg {que eu saiba que}} ela vai vir atras{\dg {!}} \\
    & \small \textsc{multi-task tag-style}  &  {\dg {Eu não}} preciso pedir {\dg {que eu saiba que}} ela vai vir atras{\dg {!}} \\
    & \small \textsc{translate train back.} & {\dg {Não}} preciso pedir {\dg {porque}} sei {\dg {que}} ela vai vir atras{\dg {!}}  \\      \addlinespace[1mm]
    
    \addlinespace[1em]
   
    \toprule[2pt]
    \rowcolor{gray!10}
    (\french)
    &  \small \textbf{\textsc{informal} }   &   drôle heinnnnnnnnn s étais ma femme de ménage!    \\ 
    \rowcolor{gray!10}
    &  \small \textbf{\textsc{xformal rewrite}} & Le plus amusant est le fait qu'il s'agissait de mon ancienne employée de maison.   \\

    \addlinespace[2mm]
    &  \small \textsc{rule-based}                     &  Drôle {\red {hein}} s étais ma femme de ménage!                                \\
    &  \small \textsc{round-trip mt}                  &  {\red {C'était}} drôle ma {\red {bonne}}!                                          \\
    &  \small \textsc{translate train tag}            & {\red { C'est}} drôle, {\red {mais j}}'étais ma femme de ménage{\red {.}}                     \\
    &  \small \textsc{multi-task tag-style}           &  {\red {C'était}} drôle, {\red {c'était}} ma femme de ménage!                        \\
    &  \small \textsc{translate train back.}          &  {\red {C'était}} drôle, {\red {mais j}}'étais ma femme de ménage!                \\ \addlinespace[1mm]
  
   \addlinespace[1em]
   
    \toprule[2pt]
    \rowcolor{gray!10}
    (\italian)
    & \small \textbf{\textsc{informal}}              &   un po'di raffreddore ma tutto ok!!!   \\ 
    \rowcolor{gray!10}
    & \small \textbf{\textsc{xformal rewrite}}       & Sono affetta da un leggero raffreddore ma per il resto va tutto bene.\\

    \addlinespace[2mm]
    & \small \textsc{rule-based}                     &  {\blue {Un}} po'di raffreddore ma tutto ok{\blue {!}}                             \\
    & \small \textsc{round-trip mt}                  &  {\blue {Un}} po'{\blue {freddo ma va bene!}}                                   \\
    & \small \textsc{translate train tag}            &  {\blue {Un}} po'di raffreddore, ma tutto {\blue {va bene.}}                   \\
    & \small \textsc{multi-task tag-style}           &  Un po'di raffreddore, ma {\blue {va bene.}}                         \\
    & \small \textsc{translate train back.}          &  Un po'di raffreddore, ma tutto {\blue {va bene.}}                      \\ \addlinespace[1mm]

    \addlinespace[0.5em]
    \toprule[2pt]
    
    \end{tabular}}
    \caption{Example system outputs on random sentences of \textsc{xformal}.}
    \label{tab:system_outputs}
\end{table*}
\begin{figure*}[ht!]
  \centering
  \begin{subfigure}[b]{0.354\linewidth}  
    \includegraphics[width=\linewidth]{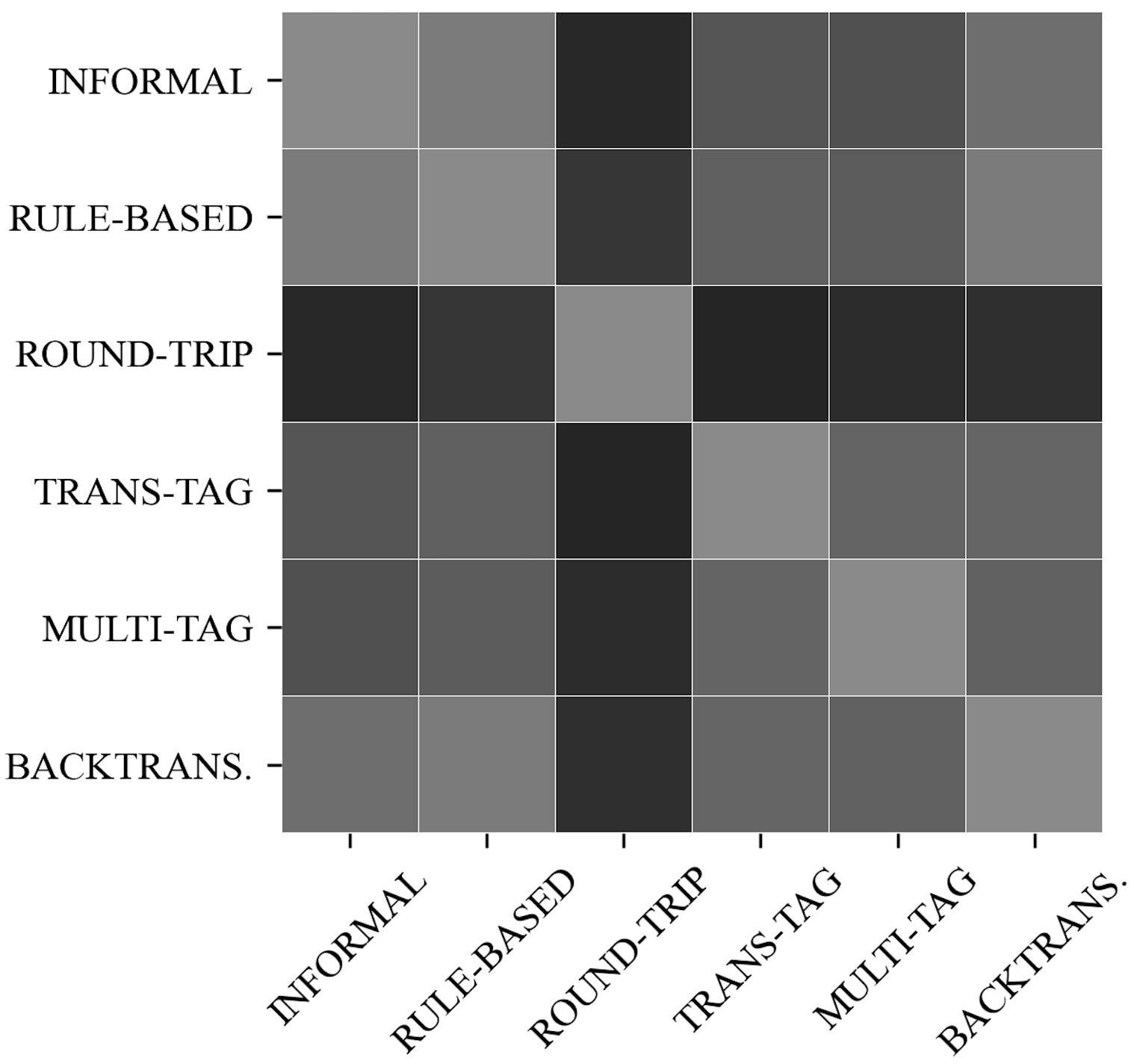}
     \caption{\portuguese}
  \end{subfigure}\hspace{0.001\textwidth}
  \begin{subfigure}[b]{0.273\linewidth}  
    \includegraphics[width=\linewidth]{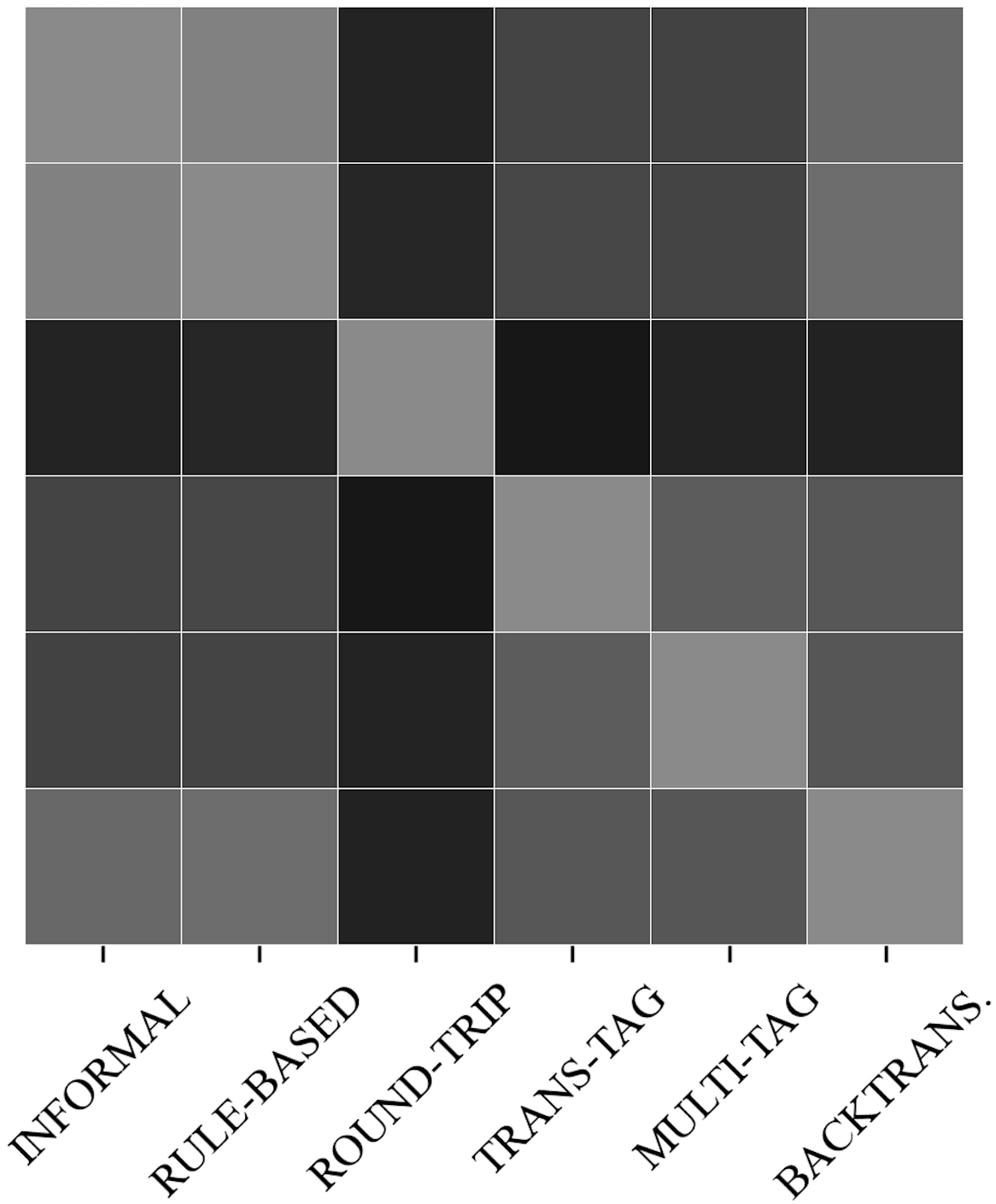}
    \caption{\french}
  \end{subfigure}\hspace{0.001\textwidth}
  \begin{subfigure}[b]{0.33\linewidth}  
    \includegraphics[width=\linewidth]{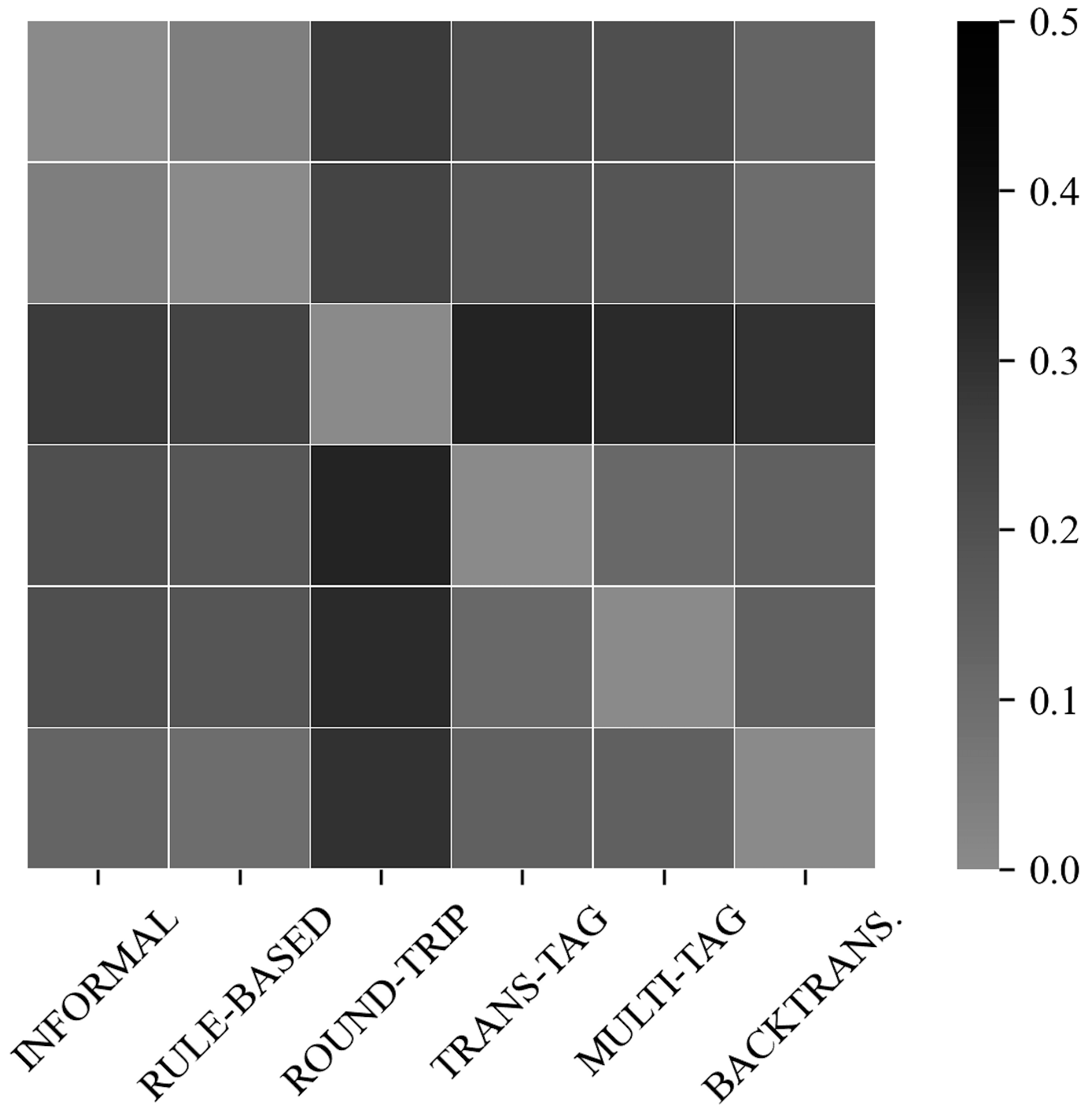}
    \caption{\italian}
  \end{subfigure}
  \caption{Heatmaps of LeD scores showing the lexical difference between pairs of systems.}
  \label{fig:led_plots}
\end{figure*}
\paragraph{Unsupervised model evaluation}\label{par:unsupervised_discussion}
We also benchmark the unsupervised models but focus solely on automatic metrics since they lag behind their supervised counterparts. As shown in Table~\ref{tab:results_all}, when using out-of-domain data (e.g., OpenSubtitles)  for training, the models perform worse than their \textsc{nmt} counterparts across all three languages.  The difference is most stark when considering self-\bleu and multi-\bleu scores.  However, given access to large \textit{in-domain} corpora (e.g., L$26$ Yahoo! French Answers) the gap between the two model classes closes with \textsc{dlsm} achieving a multi-\bleu score of 42.1 compared to 48.3 for the best performing \textsc{nmt} model \textsc{backtranslate}.  This shows the promise of unsupervised methods, assuming a large amount of in-domain data, on multilingual \st tasks.
\paragraph{Lexical differences of system outputs} Finally, in Figure~\ref{fig:led_plots} we analyze the diversity of outputs by leveraging LeD scores resulting from pair-wise comparisons of different \textsc{nmt} systems. A larger LeD score denotes a larger difference between the lexical choices of the two systems under comparison. First, we observe that the \textsc{round-trip mt} outputs have the smallest lexical overlap with the informal input sentences. However, when this observation is examined together with human evaluation results, we conclude that the large number of lexical edits happens at the cost of diverging semantically from the input sentences. Moreover, we observe that the average lexical differences within \textsc{nmt}-based systems are small. This indicates that different systems perform similar edit operations that do not deviate a lot from the input sentence in terms of their lexical choices. This is unfortunate given that multilingual \fst requires systems to perform more phrase-based operations, as shown in the analysis in \S\ref{sec:data_analysis}. 
\paragraph{Evaluation Metric}\label{par:eval_metrics}
While evaluating evaluation metrics is not a goal of this work (though the data can be used for that purpose), we observe that the top models identified by the automatic metrics generally align with the top models identified by humans.  While promising, further work is required to confirm if the automatic measures really do correlate with human judgments.

\section{Conclusions \& Future Directions}
This work extends the task of formality style transfer to a multilingual setting.  Specifically, we contribute \datasetname, an evaluation testbed consisting of informal sentences and multiple formal rewrites spanning three languages: \portuguese, \french, and \italian.  As in \citet{rao-tetreault-2018-dear} Turkers can be effective in creating high quality \st corpora.  In contrast to the aforementioned \english corpus, we find that the rewrites in \datasetname tend to be more diverse, making it a more challenging task.

Additionally, inspired by work on cross-lingual transfer and \english \fst, we benchmark several methods and perform automatic and human evaluations on their outputs.  We found that \textsc{nmt}-based ensembles are the best performing models for \french and \portuguese---a result  consistent with \english---however, they perform comparably to a naive \textsc{rule-based} baseline for \italian.  To further facilitate reproducibility of our evaluations and corpus creation processes, as well as drive future work, we will release our scripts, rule-based baselines, source data, and annotation templates, on top of the release of \datasetname.  

Our results open several avenues for future work in terms of benchmarking and evaluating \fst in a more language inclusive direction. Notably, current supervised \textbf{and} unsupervised approaches for \english \fst rely on parallel in-domain data---with the latter treating the parallel set as two unpaired corpora---that are not available in most languages. We suggest that \textbf{benchmarking} \fst models in multilingual settings will help understand their ability to generalize and lead to safer conclusions when comparing approaches.
At the same time, multilingual \fst calls for more language-inclusive consideration for \textbf{automatic evaluation} metrics. Model-based approaches have been recently proposed for evaluating different aspects of \st. However, most of them rely heavily on English resources or pre-trained models. How those methods can be extended to multilingual settings and how we evaluate their performance remain open questions.

\section*{Acknowledgements}

We thank 
Sudha Rao for providing references and materials of the \textsc{gyafc} dataset,
Chris Callison-Burch and Courtney Napoles for discussions on MTurk annotations,
Svebor Karaman for helping with data collection, our colleagues at Dataminr, and the
\textsc{naacl} reviewers for their helpful and constructive comments.


\section{Ethical Considerations}

Finally, we address ethical considerations for \textbf{dataset papers} given that our work proposes a new corpus \datasetname.  We reply to the relevant questions posed in the \textsc{naacl} $2021$ Ethics \textsc{faq}.\footnote{\url{https://2021.naacl.org/ethics/faq/}}

\subsection{Dataset Rights}

The underlying data for our dataset as well as training our \fst models and formality classifiers are from Yahoo! Answers L6 dataset.  We were granted written permission by Yahoo (now Verizon) to make the resulting dataset public for academic use.

\subsection{Dataset Collection Process}

Turkers are paid over $10$ \textsc{usd} an hour.  We targeted a rate higher than the \textsc{us} national minimum wage of $7.50$ \textsc{usd} given discussions with other researchers who use crowdsourcing.
We include more information on collection procedures in~\S\ref{sec:data_collection}.

\subsection{IRB Approval}

This question is not applicable for our work.

\subsection{Dataset Characteristics}

We follow \citet{bender-friedman-2018-data} and \citet{datasheets} and report characteristics in \S\ref{sec:data_collection} and \S\ref{sec:data_analysis}.\label{sec:ethics}

\bibliography{naacl2021}
\bibliographystyle{acl_natbib}

\appendix
\clearpage
\section{Does translation preserve formality?}\label{sec:formality_artifacts}
We examine the extend to which machine translation---through the \textsc{aws} service---affects the formality level of an input sentence: starting from a set of English sentences we have formality judgments for ({\fontfamily{qcr}\selectfont ORIGINAL-\textbf{EN}}), we perform a round-trip translation via pivoting through an auxiliary language ({\fontfamily{qcr}\selectfont PIVOT-\textbf{X}}). We then compare the formality prediction scores of the English Formality regression model for the two versions of the English input. In terms of Spearman correlation, the model's performance drops by $7.5$ points on average when tested on round-trip translations. To better understand what causes this drop in performance, we present a per formality bin analysis in Table \ref{tab:round_trip_experiment}. On average we observe that translation preserves the formality level of formal sentences considerably well, while at the same time it tends to shift the formality level of informal sentences towards formal values---by a margin smaller than $1$ point---most of the times. 
To account for the formalization effect of translation, we draw the line between formal and informal sentences at the value of $-1$ for scores predicted by multilingual regression models. This decision is based on the following intuition:
if the formality shift of machine translated informal sentences is around $+1$ value, the propagation of English formality labels imposes a negative shift of formal sentences in the model's predictions.
\begin{table*}[!ht]
    \centering
    \scalebox{0.75}{
    \begin{tabular}{l@{\hskip 0.4in}rrr@{\hskip 0.4in}rrr}
       & \multicolumn{3}{c}{\textit{Informal bins}} & \multicolumn{3}{c}{\textit{Formal bins}}\\
       & [-3,-2]        & [-2,-1]            & [-1,0]       & [0,1]           & [1,2]             & [2,3]         \\
        \toprule[1.5pt]
    {\fontfamily{qcr}\selectfont GOLD-STANDARD}     & $-2.35 \pm 0.31$ &  $-1.49 \pm 0.36$  & $-0.57 \pm 0.33$   & $0.46 \pm 0.32$ & $1.33 \pm 0.34$   & $2.27 \pm 0.25$\\
        \toprule[1pt]
    {\fontfamily{qcr}\selectfont ORIGINAL-\textbf{EN}} & $-1.81 \pm 0.59$ &  $-1.18 \pm 0.73$  & $-0.53 \pm 0.71$   & $0.30 \pm 0.82$ & $1.17 \pm 0.68$   & $1.50 \pm 0.45$\\
    \toprule[0.02pt]
    \addlinespace[0.2cm]
    {\fontfamily{qcr}\selectfont PIVOT-\textbf{IT}} & $-1.58 \pm 0.67$ & $-0.95 \pm 0.74$   & $-0.35 \pm 0.68$    & $0.34 \pm 0.86$ & $1.12 \pm 0.73$   & $1.41 \pm 0.54$\\
    $\Delta$ & $\color{blue}{\mathbf{+0.23 \pm 0.45}^*}$ & $\color{blue}\mathbf{+0.23 \pm 0.54^*}$   & $\color{blue}\mathbf{+0.18 \pm 0.53^*}$    & $+0.04 \pm 0.39$ & $\color{red}\mathbf{-0.05 \pm	0.23}$   & $\color{red}\mathbf{-0.09 \pm	0.24}$\\
    \addlinespace[0.2cm]

    {\fontfamily{qcr}\selectfont PIVOT-\textbf{FR}} & $-1.44 \pm 0.77$ & $-0.93 \pm 0.75$  & $-0.37 \pm 0.72$    & $0.38 \pm 0.80$   & $1.12 \pm 0.82$   & $1.50 \pm 0.47$\\
    $\Delta$ & $\color{blue}\mathbf{+0.37 \pm 0.54^*}$ & $\color{blue}\mathbf{+0.25 \pm 0.50^*}$  & $\color{blue}\mathbf{+0.15 \pm 0.48}$    & $\color{blue}\mathbf{+0.08 \pm 0.42}$  & $\color{red}\mathbf{-0.05 \pm 0.49}$   & $-0.00 \pm 0.17$ \\

    \addlinespace[0.2cm]
    {\fontfamily{qcr}\selectfont PIVOT-\textbf{PT}} & $-1.46 \pm 0.71$ & $-0.87 \pm 0.80$  & $-0.31 \pm 0.70$    & $0.34 \pm 0.83$   & $1.14 \pm 0.71$   & $1.47 \pm 0.47$ \\
    $\Delta$ & $\color{blue}\mathbf{+0.35 \pm 0.49^*}$ & $\color{blue}\mathbf{+0.31 \pm 0.57^*}$  & $\color{blue}\mathbf{+0.21 \pm 0.55^*}$    & $+0.03 \pm 0.39$  & $-0.03 \pm 0.31$   & $-0.03 \pm 0.14$ \\
    \toprule[1.5pt]
    \end{tabular}}
    \caption{Average formality scores of human annotations ({\fontfamily{qcr}\selectfont GOLD-STANDARD}) and model's predictions on the original and round-trip translated PT16 test set grouped in $6$ bins of varying formality.  
    $\Delta$ gives the average formality shift of the English sentences resulting from round-trip translation. Scores in blue and red indicate that mean is above and below zero, respectively. $*$ denotes statistical significant formality shifts with $p<0.05$. Translation preserves formality of formal sentences while informal sentences exhibit a shift towards formal values.}
    \label{tab:round_trip_experiment}
\end{table*}
\section{Amazon Web Service details}\label{sec:aws_details}
We compute the performance of the \textsc{aws} system on $2.5$K randomly  sentences from OpenSubtitles, as a sanity check of translation performance. We report \bleu of  $37.16$ (\textsc{br-pt}),  $33.79$ (\textsc{fr}), and $32.67$ (\textsc{it}).
\section{Rule-based baselines}\label{sec:rules_details}
We develop a set of rules to automatically make an informal sentence more formal via performing surface-level edits 
similar to the \english rule-based system of \citet{rao-tetreault-2018-dear}. The set of extracted rules 
are shared across languages with the only difference being the list of abbreviations:
\paragraph{Normalize punctuation} We remove punctuation symbols that are repeated $>=2$ times. 
\paragraph{Character repetition} We trailed characters repeated $>=3$ times (\textit{e.g., ciaooo} $\rightarrow$ \textit{ciao} (\italian)).
\paragraph{Normalize casing} Several sentences might consist of words that are written in upper case. We lower case all characters apart from the first characters of the first word. 
\paragraph{Normalize abbreviations} Informal text might contain slang words that are be abrreviated. We hand-craft a list of expansions for each language. For example, we replace \textit{kra} $\rightarrow$ \textit{cara} (\portuguese), \textit{ta} $\rightarrow$ \textit{ti amo} (\italian), and \textit{bjr} $\rightarrow$ \textit{bonjour} (\french), using publicly available resources. \footnote{\url{https://braziliangringo.com/brazilianinternetslang/}}\textsuperscript{,}
\footnote{\url{https://www.dummies.com/languages/italian/texting-and-chatting-in-italian/}}\textsuperscript{,}
\footnote{\url{https://frenchtogether.com/french-texting-slang/}}
The resulting lists sizes are: $64$ (\portuguese), $21$ (\italian), and $48$ (\french).
\section{NMT architecture details}\label{sec:nmt_details}
We used the \textsc{nmt} implementations of \citet{niu-etal-2018-multi}  that are publicly available: \url{https://github.com/xingniu/multitask-ft-fsmt}.
The \textsc{nmt} models are implemented as bi-directional \textsc{lstm}s on Sockeye \cite{hieber-etal-2018-sockeye}, using the same configurations across languages to establish fair comparisons. We use single \textsc{lstm}s consisting of a single of size $512$, multilayer perceptron attention with a layer size of $512$, and word representations of size $512$. 
We apply layer normalization and tie the source and target embeddings as well as the output layer’s weight matrix. We add dropout with probability $0.2$ (for the embeddings and \textsc{lsmt} cells in both the encoder and the decoder). For training, we use the Adam optimizer with a batch size of $64$ sentences and checkpoint the model every $1000$ updates. Training stops after $8$ checkpoints without improvement of validation perplexity. For decoding, we use a beam size of $5$.
\section{Inter-annotator agreement on human evaluation}\label{sec:human_agreement}
To quantify inter-annotator agreement for the tasks of \textit{formality}, \textit{meaning preservation}, and \textit{fluency} we measure the correlation of their ordinal ratings using inter-class correlation (\textsc{icc}) and their categorical agreement using a variations of Cohen's $\kappa$ coefficient. For the latter, given that we collect human evaluation judgments through crowd-sourcing, we follow the simulation framework of \citet{pavlick-tetreault-2016-empirical} to quantify agreement. Concretely, we simulate two annotators (\textit{Annotator} $1$, \textit{Annotator} $2$) via randomly choosing one annotator's judgment for a given instance as the rating of \textit{Annotator $1$} and taking the mean rating of the rest judgments as the rating of \textit{Annotator $2$}. We then compute Cohen's $\kappa$ for these two simulated annotators. We repeat this process $1{,}000$ times, and report the median and standard deviation of results. For measuring agreement of the \textit{overall} ranking evaluation task we use the same simulated framework and report results of Kendall's $\tau$. Table~\ref{tab:human_agreement} presents Inter-annotator agreement results on human evaluation across evaluation aspects and languages. 
\begin{table*}[!ht]
    \centering
    \scalebox{0.8}{
    
    \begin{tabular}{l@{\hskip 0.5in}cc@{\hskip 0.5in}cc@{\hskip 0.5in}cc}
    
    \toprule[1.5pt]
    \textbf{\textsc{task}} & \multicolumn{2}{c}{\textsc{br-pt}} & \multicolumn{2}{c}{\textsc{fr}} & \multicolumn{2}{c}{\textsc{it}} \\ 
    \addlinespace[0.1cm]
    \toprule[0.02pt]
    \addlinespace[0.15cm]
    \addlinespace[1em]

                         &  Weighted $\kappa$ &  ICC  &    Weighted $\kappa$ &  ICC  &   Weighted $\kappa$ &  ICC            \\ 
    \addlinespace[1em]
    Formality            &  $0.57 \pm 0.02$  & $0.72 \pm 0.01$   & $0.56 \pm 0.02$ & $0.71 \pm 0.01$ & $0.67 \pm 0.02$ & $0.79 \pm 0.02$\\  
    Fluency              &  $0.43 \pm 0.02$  & $0.62 \pm 0.02$   & $0.58 \pm 0.02$ & $0.73 \pm 0.01$ & $0.46 \pm 0.02$ & $0.64 \pm	0.02$\\
    Meaning Preservation &  $0.53 \pm 0.02$  & $0.70 \pm 0.02$   & $0.71 \pm 0.02$ & $0.83 \pm 0.01$ & $0.55 \pm 0.03$ & $0.71 \pm 0.02$ \\
    \addlinespace[1em]
    
                         &  \multicolumn{6}{c}{Kendall's $\tau$} \\ 
    \addlinespace[1em]
    Overall              & \multicolumn{2}{c}{$0.41 \pm 0.03$} & \multicolumn{2}{c}{$0.44 \pm 0.03$} & \multicolumn{2}{c}{$0.41 \pm 0.03$}  \\
    \addlinespace[0.5em]
    \toprule[1.5pt]
    \end{tabular}}
    \caption{Inter-annotator agreement for human evaluation results.}
    \label{tab:human_agreement}
\end{table*}
\section{Yahoo! L6 language statistics}\label{sec:multi_stats_l6}
Table~\ref{tab:l6_multi_stats} presents the total number of questions included in L$6$ Yahoo! Corpus, for each language. Although almost $90\%$ of questions are in English, the corpus contains a non-neglibible number of informal sentences in Spanish, French, Portuguese, Italian, and German, with a long tail of few questions for other languages.
\begin{table}[!ht]
    \centering
    \scalebox{0.7}{
    \begin{tabular}{l@{\hskip 0.6in}c@{\hskip 0.8in}r}
    \textsc{language} & \textsc{code} & $\#$ \textsc{questions}\\
    \hline
    English & \textsc{en} & $3{,}895{,}407$ \\ 
    Spanish & \textsc{es} & $258{,}086$ \\
    French & \textsc{fr} & $125{,}393$ \\
    Portuguese & \textsc{pt} & $105{,}813$ \\
    Italian & \textsc{it} & $55{,}027$ \\
    German & \textsc{de} & $43{,}149$ \\
    Persian & \textsc{fa} & $30$ \\
    Catalan & \textsc{ca} & $27$ \\
    Dutch & \textsc{nl} & $13$ \\ 
    Danish & \textsc{da} & $10$ \\
    Arabic & \textsc{ar} & $8$ \\
    Romania & \textsc{ro} & $8$ \\
    Norwegian & \textsc{no} & $8$ \\
    Swedish & \textsc{sv} & $5$ \\ 
    Estonian &  \textsc{et} & $5$ \\
    Finnish & \textsc{fi} & $4$ \\
    Malay & \textsc{ms} & $4$ \\
    Turkish & \textsc{tr} & $4$ \\
    Slovak & \textsc{sk} & $3$ \\
    Latvian & \textsc{lv} & $3$ \\ 
    Albania & \textsc{sq} & $3$ \\
    Croatian & \textsc{hr} & $3$ \\
    Tagalog & \textsc{tl} & $3$ \\
    Chech & \textsc{cs} & $3$ \\
    Slovenian &  \textsc{sl} & $3$ \\
    Vietnamese &  \textsc{vi} & $2$\\
    Icelandic &  \textsc{is} & $2$ \\
    Hungarian &  \textsc{hu} & $2$ \\
    Polish & \textsc{pl} & $2$ \\
    \end{tabular}}
    \caption{Number of question per language included in L$6$ Yahoo! Answers.}
    \label{tab:l6_multi_stats}
\end{table}
\section{Instructions for \datasetname}\label{sec:intructions_for_rewrites}
\paragraph{Summary of the task} Given an \textbf{informal sentence} in French (or Portuguese/Italian), generate its \textbf{formal rewrite}, 
\textit{without changing its meaning}.
\paragraph{Detailed Instructions} Given an informal sentence, provide us with its informal rewrite. The informal rewrite should only change the formality attribute of the original sentence and preserve its meaning. Each sentence should be treated independently while rewrites should only rely on the information available in the sentences. There is no need to guess what additional information might be available in the documents the sentences come from.
\paragraph{Examples} Following we include examples of good and bad rewrites given to Turkers: 
\begin{tabular}{p{0.9\columnwidth}}
\\
\textsc{informal} \it Wow, I am very dumb in my observation skills......\\
\textsc{good formal} \it I do not have good observation skills.\\
\textcolor{darkgray}{\textsc{reasoning} \it  Formality is properly transferred and meaning is preserved.}\\
\\
\textsc{bad formal} \it Wow, I am very foolish in my observation skills..\\
\textcolor{darkgray}{\textsc{reasoning} \it  Formality is not properly transferred.}\\
\\
\textsc{bad formal} \it I am very unintelligent and I don't have good observation skills.\\
\textcolor{darkgray}{\textsc{reasoning} \it  Meaning has changed.}\\
\end{tabular}
\section{Qualitative analysis of \datasetname}\label{sec:qualitative_analysis_details}
Following, we include the instructions given to \textbf{native speakers} for the qualitative analysis of \datasetname, as described in \S\ref{sec:data_analysis}.
\paragraph{Background context} In this task you will be asked to judge the quality of formal rewrites of a set of informal sentences. To give more background context, your work would serve as a quality check over annotations obtained from the Amazon Mechanical Turk (AMT) platform.  AMT  workers were given an informal sentence, e.g.,  ``I’d say it is punk though'', and then asked to provide us with its formal rewrite while maintaining the original content and being grammatical/fluent, e.g., ``However, I do believe it to be punk''.
In our work, workers were presented with sentences in French, Italian and Brazilian Portuguese. For our analysis we want to know a) whether the quality of the collected annotations is good (Task 1), b) what are the types of edits workers performed when formalizing the input sentence (Task 2).    
Both tasks consist of the same $200$ informal-formal sentences-pairs and could be performed either in parallel (e.g., judging a single informal-formal sentence-pair both in terms of quality and types of edit at the same time; Task 1 and 2 in Google sheet), or individually (Task 1, Task 2 in Google sheet). More information and examples for the two tasks are included below. 
\paragraph{Task 1} Given an informal sentence you are asked to assess the quality of its formal rewrite. For each sentence pair type excellent, acceptable, or poor under the rate column.  
Read the instructions below before performing the task!\\
\noindent
\textbf{What constitutes a good formal rewrite?}
\begin{itemize}
    \item The style of the rewrite should be formal
    \item The overall meaning of the informal sentence should be preserved
    \item The rewrite should be fluent 
\end{itemize}
\noindent
\textbf{How should I interpret the provided options?} 
Below we include detailed instructions on how to interpret the provided options. 
\begin{itemize}
    \item \textbf{Excellent} the rewrite is formal, fluent and the original meaning is maintained.  There is very little that could be done to make a better rewrite.
    \item \textbf{Acceptable} the rewrite is generally an improvement upon the original but there are some minor issues (such as the rewrite contains a typo or missed transforming some informal parts of the sentence, for example).  
    \item \textbf{Poor} the rewrite offers a marginal or no improvement over the original.  There are many aspects that could be improved.
\end{itemize}
\paragraph{Task 2} 
In this task the goal is to characterize the types of edits workers made while formalizing the informal input sentence. For each informal-formal sentence pair you should check each of the provided boxes. Note that: a) 
Multiple types of edits might hold at the same type (e.g., in the informal---formal  ``Lexus cars are awesome!''---``Lexus cars are very nice.'' you should check both the paraphrase and the punctuation boxes.); b)
At the end of the Google sheet there is a column named ‘Other’ you are welcome to write down in plain text any additional type that you observed and it is not covered by the existing classes. The provided classes are: capitalization,	punctuation, paraphrase, delete fillers, completion, add context, contractions, spelling, normalization, slang/idioms,	politeness, split sentences, and relativizers.
\section{Quantitative analysis of \datasetname}\label{sec:quantitative_analysis_details}
Following, we include more details on the qualitative analysis procedures.
\paragraph{Capitalization} A rewrite performs a capitalization edit if it contains tokens that appear in capital letters in the informal text but in lowercase letters in the formal text.
\paragraph{Punctuation} A rewrite contains a punctuation edits if any of the punctuation of the informal-formal texts differs. 
\paragraph{Spelling} We identify spelling errors based on the character-based Levenshtein distance between informal-formal tokens. 
\paragraph{Normalization} We identify normalization edits based on a hand-crafted list of abbreviations for each language. 
\paragraph{Split sentences} We split sentences using the \textsc{nltk} toolkit. 
\paragraph{Paraphrase} A formal rewrite is considered to contain a paraphrase edit if the token level Levenshtein distance between the informal-formal text is greater than 3 tokens.
\paragraph{Lowercase} A rewrite performs a capitalization edit if it contains tokens that appear in lower case in the informal text but in capital letters in the formal text.
\paragraph{Repetition} We identify repetition tokens using regular expressions (a token that appears more than 3 times in a row is considered a repetition.)
\section{\datasetname: Data Quality}\label{sec:data_quality}
We ask a \textbf{native speaker} to judge the quality of a sample of $200$ rewrites for each language via choosing one of the following three options: ``excellent'', ``acceptable'', and ``poor''. The details of this analysis are included in~\appendixmacro\ref{sec:qualitative_analysis_details}~ (\textbf{Task 1}).Results indicate that on average less than $10\%$ of the rewrites were identified as of poor quality across languages while more than $40\%$ as of excellent quality. The small number of rewrites identified as ``poor'' consists mostly of edits where humans add context not appearing in the original sentence. We choose not to exclude any of the rewrites from the dataset as we provide multiple reformulations of each informal instance. 
\section{OpenSubtitles data}\label{sec:opensubs_extra_stats}
\begin{table}[!ht]
    \centering
    \scalebox{0.8}{
    \begin{tabular}{lcrrr}
    OpeSubtiles & Preprocessing & \portuguese & \italian & \french  \\
    \multirow{2}{*}{All}& \xmark & $61.3$M & $35.2$M & $41.8$M\\
    & \cmark & $29.8$M & $16.8$M & $19.0$M\\
    Informal & \cmark & $25.7$M & $14.7$M & $16.3$M\\
    \end{tabular}}
    \caption{OpenSubtitles statistics in three languages.}
    \label{tab:opensubs_statistics}
\end{table}
\section{Qualification tests}\label{sec:qualification_tests}
Table~\ref{tab:qualification_test_all} presents questions and answers used in \textbf{\textsc{qc2}} of our annotation protocol. Turkers have to score $80$ and above to participate in the task. To compute an average score for each test, we assume that an answer is incorrect if it deviates more than $1$ point from the gold-standard scores given in the second column of Table~\ref{tab:qualification_test_all}.
\begin{figure*}[ht!]
  \centering
  \begin{subfigure}[b]{0.24\linewidth}
    \includegraphics[width=\linewidth]{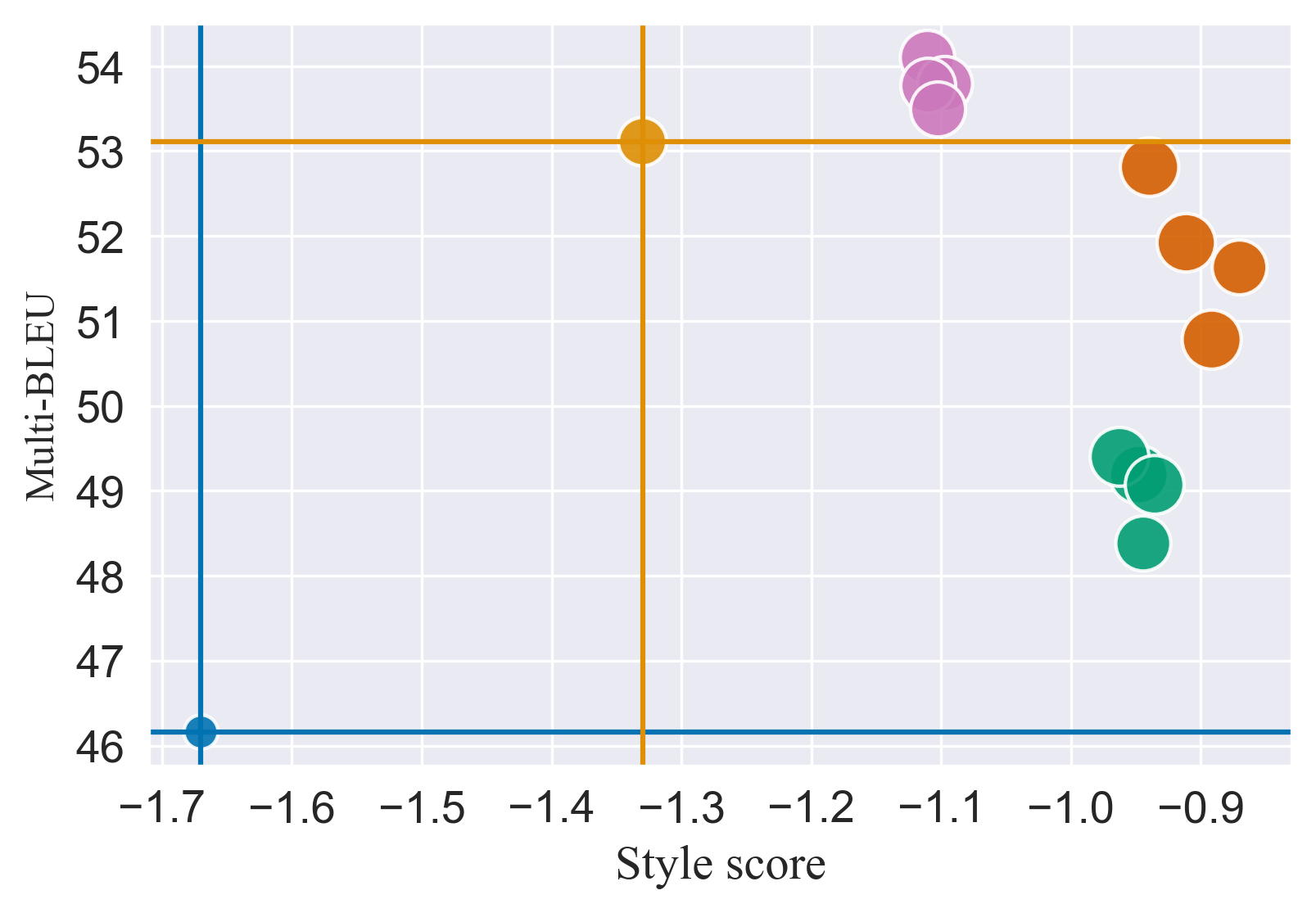}
     \caption{\portuguese}
  \end{subfigure}
  \begin{subfigure}[b]{0.23\linewidth}
    \includegraphics[width=\linewidth]{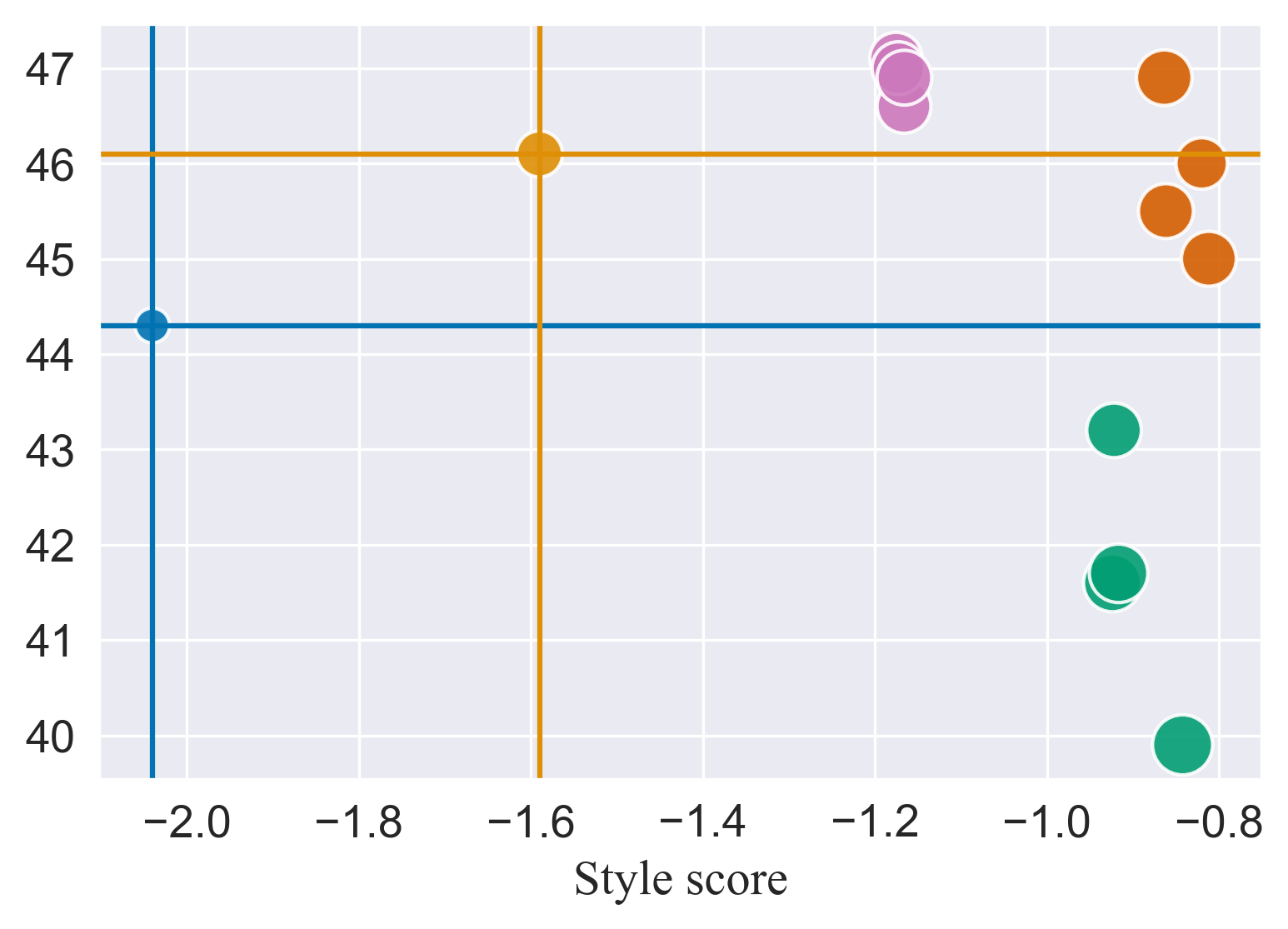}
    \caption{\french}
  \end{subfigure}
  \begin{subfigure}[b]{0.23\linewidth}
    \includegraphics[width=\linewidth]{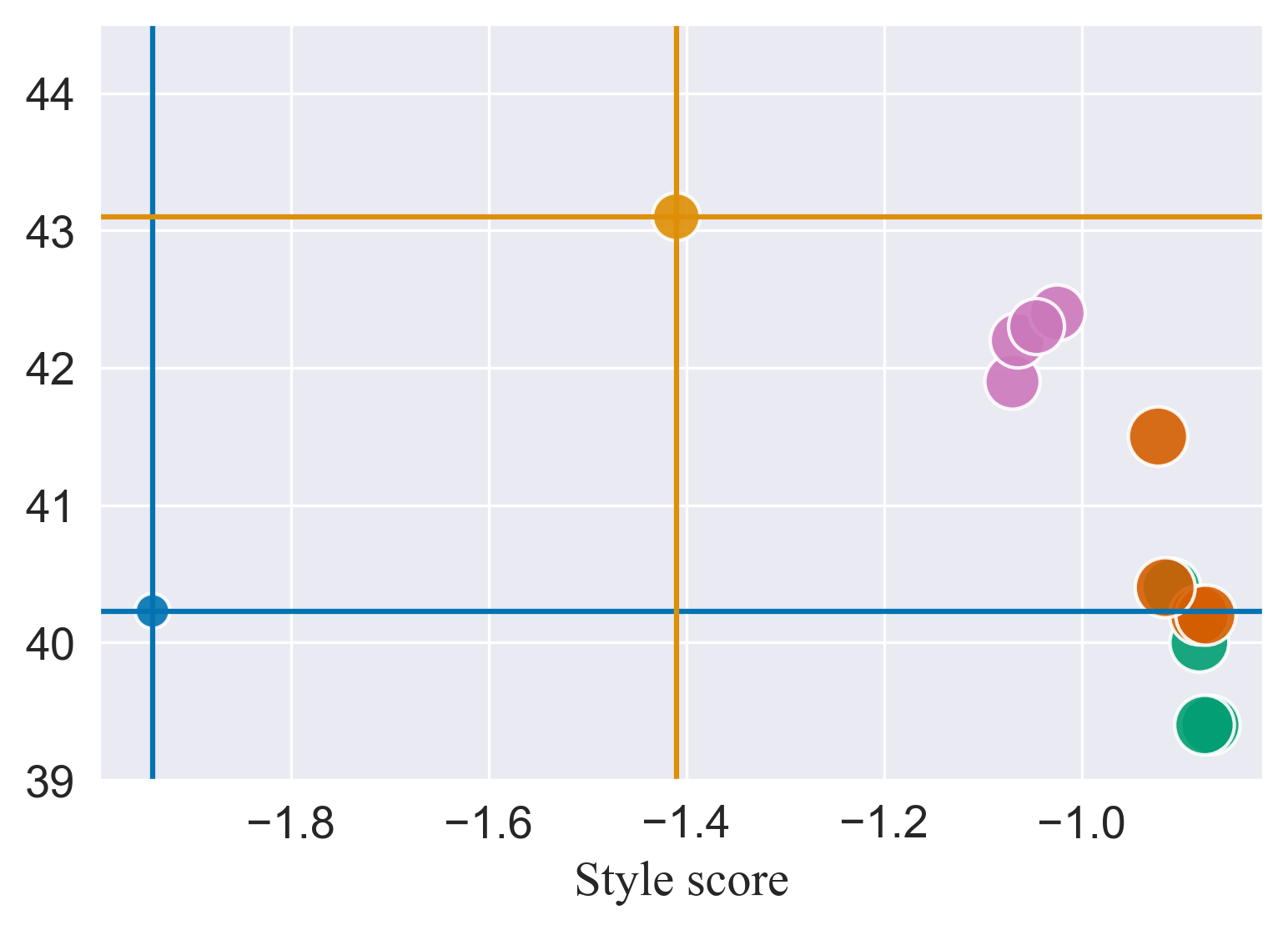}
    \caption{\italian}
  \end{subfigure}
    \begin{subfigure}[b]{0.11\linewidth}
    \includegraphics[width=\linewidth]{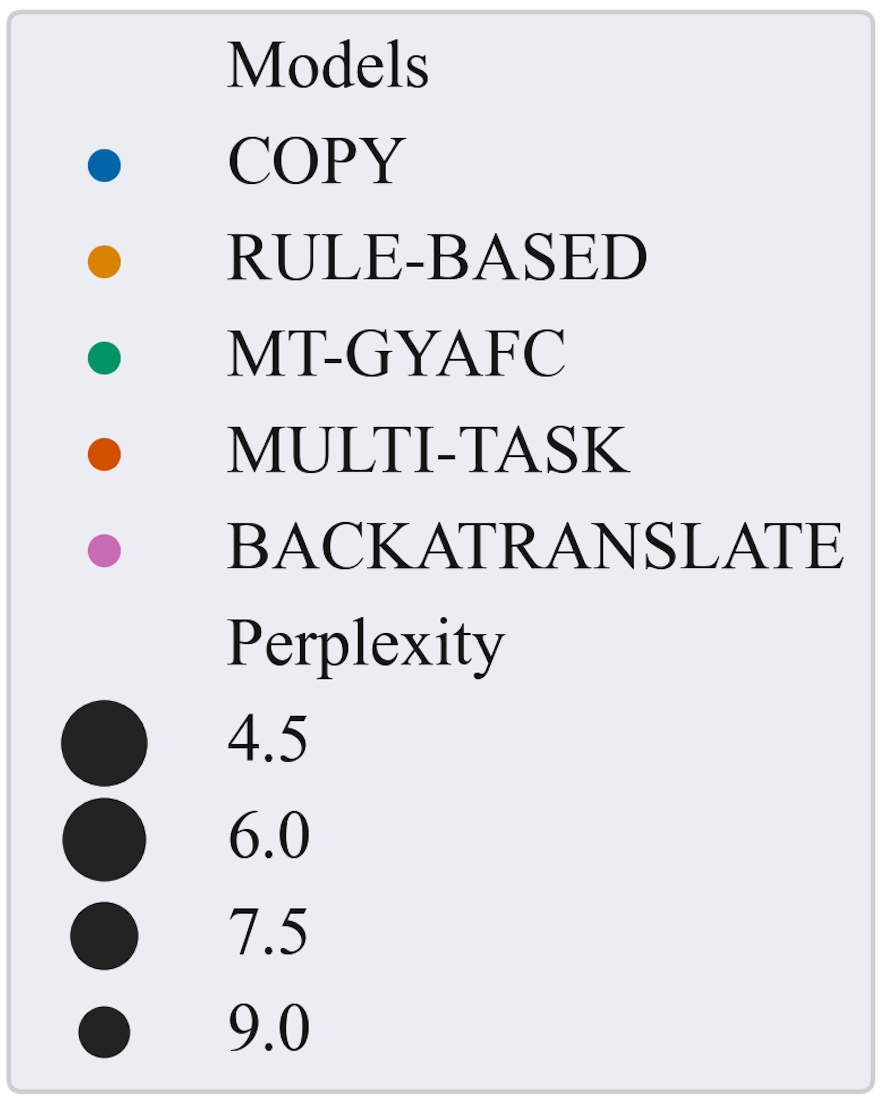}
    \vspace{1.3em}
  \end{subfigure}
  \caption{\textbf{\bleu vs.\ \textsc{style score} vs.\ \textsc{perplexity}} trade-off plots across $4$ reruns.}
  \label{fig:tradeoff_blue_vs_style_score}
\end{figure*}
\begin{table*}[ht!]
    \centering
    \scalebox{0.77}{
    \begin{tabular}{lr}
    
        \hline
        \textbf{\textsc{french}} &  \\
        \hline
        \addlinespace[1em]
        Heureux ceux qui ont la chance de pouvoir faire la sieste devant Derrick ! & $-2$ \\
        je kiffe trop ce film il déchire carément!! & $-3$\\
        Moi je les données à des vieilles personnes qui ne comprennent pas le fonctionnement d'un lecteur DVD! & $-1$\\
        Le producteur de ce morceau, c'est Scott Torch. & $2$\\
        Il s'agit bien du groupe America et le titre "A horse with no name". & $2$\\
        La France ne gagnera jamais si elle continue avec des chansons qui ont l'air de dater des années 70. & $0$ \\
        Le but n'est pas de se faire du profit dessus! & -1 \\ 
        Il existe surtout sur les chaînes du câble et du satellite, TL9, TPS foot, par exemple. & $2$\\
        
        \addlinespace[1em]
        \hline

        \textbf{\textsc{italian}} &  \\
        \hline
        \addlinespace[1em]
        
        TI do questo sito qui puoi trovare qualunque cosa su Tru Calling buona lettura! & $-1$ \\
        Parla proprio di amore in chat. & $-2$ \\
        ma parla cm mangi!!!!!!!!!!! &  $-3$ \\
        io ho un programma si kiama evil lyrics.  & $-1$ \\
        Se non ci si dà una svegliata "dove andremo a finire?" & $-1$ \\
        Alla fine del Festival ci saranno due vincitori, uno per categoria. & $2$ \\
        è quello che stò passando ora. & $-1$ \\
        Il montaggio risulta quindi ridotto al minimo. & $2$ \\
        
        \addlinespace[1em]
        \hline
        \textbf{\textsc{brazilian-portuguese}} &  \\
        \hline
        \addlinespace[1em]
        
        Me parece que você usa BASTANTE essa palavra e seus derivados, né?!!    & $-2$ \\
        Mas vem cá, você não tem nada melhor pra fazer do que escutar RBD ao contrário?! & $-2$ \\
        Então na minha opinião é babaquice daquele que tem preconceitos desse tipo. & $-1$ \\
        Vi a propaganda na locadora,  na minha cidade. & $0$ \\
        Na minha opinião, o mínimo existencial subentende a palavra oportunidade. & $1$ \\
        A única diferença dessa mulher para um assassino comum, é que ela nem chega a conhecer sua vítima. & $1$ \\
        Dentre as hipóteses menos consideradas por historiadores sérios está o famoso mito celta. & $1$ \\
        Dado um ambiente qualquer, uma sala ou um quarto, por exemplo, as ondas sonoras "respondem" de formas diferentes. & $-1$ \\

        \addlinespace[0.5em]
        \hline
    \end{tabular}}
    \caption{Qualification test (questions and answers) for \textsc{\textbf{qc2}}}
    \label{tab:qualification_test_all}
\end{table*}
\section{Trade-off plots}\label{sec:trade_offs_}
\paragraph{Trade-off plots} Figure~\ref{fig:tradeoff_blue_vs_style_score} presents trade-off plots between multi-\bleu vs. formality score and fluency, across different reruns as proposed by \citet{tikhonov-yamshchikov-2018-sounds}. First, we observe that models exhibit small variations in terms of formality and fluency scores across different reruns, and larger variations across \bleu for most cases. Notably, the single seed \textsc{backtranslate} systems are the most consistent across reruns for all metrics and languages. Furthermore, in almost all cases, models trained on $2$M data, perform better that the naive \textsc{copy} baseline, across metrics. However, single seed models fail to consistently outperform the \textsc{rule-based} baselines in almost all cases, with the exception of \textsc{backtranslate} for \portuguese and \french which report an improvement of  about $1$ \bleu score.
\section{Compute time $\&$ Infrastracture}
All experiments for benchmarking both \textsc{nmt}-based and unsupervised approaches use Amazon \textsc{ec}$2$ P$3$  instances: \url{https://aws.amazon.com/ec2/instance-types/p3/}, on Tesla V$100$ \textsc{gpu}s. Concretely, \textsc{nmt}-based experiments are run on $4$ \textsc{gpus} ($\sim$ $1$-$2$ hours), while unsupervised models are run on single \textsc{gpu}s, with average run time spanning from $\sim 5$ hours (for out-of-domain data---$100$K), to $\sim 72$ hours (for in domain data---$2$M).\label{sec:appendix}

\end{document}